\documentclass[preprint,review,12pt,authoryear]{elsarticle}
\usepackage[a4paper,margin=1in]{geometry}
\journal{Neural Networks}
\biboptions{authoryear}
\usepackage[hyphens]{url}
\usepackage{graphicx}
\usepackage{amsmath,amssymb,amsthm}
\usepackage{booktabs}
\usepackage{array,tabularx}
\usepackage{multirow}
\usepackage{algorithm}
\usepackage{float}
\usepackage{algpseudocode}
\usepackage{xcolor}
\usepackage{tikz}
\usepackage{pgfplots}
\pgfplotsset{compat=1.18}
\usetikzlibrary{positioning, arrows.meta, shapes.geometric, calc, fit, backgrounds}
\newtheorem{theorem}{Theorem}
\newtheorem{proposition}{Proposition}
\newtheorem{corollary}{Corollary}
\newtheorem{lemma}{Lemma}
\newtheorem{definition}{Definition}
\theoremstyle{remark}
\newtheorem{remark}{Remark}
\usepackage[unicode=true,bookmarksnumbered=true,colorlinks=true,linkcolor=blue,citecolor=blue,urlcolor=blue]{hyperref}
\newcommand{\vinj}{\hat v_{\mathrm{inj}}}
\newcommand{\Pval}{\mathcal P_{\mathrm{val}}}

\begin{document}
\begin{frontmatter}
\title{TANSAR: Editable and Attributable Correlation Matrix Memories under Correlated Keys, with Rule-Constrained Multi-Hop Knowledge Editing}

\author[ukh]{Abdulhady Abas Abdulla}
\ead{abdulhady.abas@ukh.edu.krd}
\author[elte]{Jiyan Salim Mahmud}
\ead{jiyan@inf.elte.hu}
\author[uok]{Fatemeh Daneshfar\corref{cor1}}
\ead{f.daneshfar@uok.ac.ir}
\author[tiu]{Hussein Mohammed Ali}
\ead{hussein.mohammed@tiu.edu.iq}
\author[oulu]{Mourad Oussalah}
\ead{Mourad.Oussalah@oulu.fi}
\author[lancaster]{Muhammad Bilal}
\ead{m.bilal@ieee.org}
\cortext[cor1]{Corresponding author}
\address[ukh]{University of Kurdistan Hewl\^{e}r, Erbil, Kurdistan Region, Iraq}
\address[elte]{Faculty of Informatics, E\"otv\"os Lor\'and University, Budapest, Hungary}
\address[uok]{Department of Computer Engineering, University of Kurdistan, Sanandaj, Iran}
\address[tiu]{Cybersecurity Department, Tishk International University, Erbil, Kurdistan Region, Iraq}
\address[oulu]{University of Oulu, Oulu, Finland}
\address[lancaster]{Lancaster University, Lancaster, United Kingdom}

\begin{abstract}
Correlation Matrix Memories (CMMs) store associations as sums of rank-one outer products, providing closed-form addressability, exact deletion, and linear attribution. While classical analyses assume random, orthogonal keys, semantic encoders produce correlated representations. We analyse CMMs under correlated keys using frame-theoretic coherence: (i) capacity bounds yielding an encoder-independent limit $n\le d(1+\varepsilon^2)$ for mean squared crosstalk $\varepsilon^2$; (ii) an exact characterisation of the collateral effect of deleting an association, guaranteeing that deletion never invalidates certified associations; and (iii) a geometric transparency bound demonstrating why attribution requires candidate sets of centred keys. On this basis, we introduce Topology-Aware Neuro-Symbolic Associative Retrieval (TANSAR). TANSAR traverses a directed acyclic association graph, prunes invalid paths via sound ground Horn constraints, and injects a representation derived strictly from the verified path into the residual stream of a frozen language model, guarded by a Clopper--Pearson selective-risk gate. Controlled experiments confirm all theoretical bounds. In multi-hop knowledge editing, Horn pruning reduces answer contradictions from $25.1\%$ to $0.5\%$, whereas Hopfield retrieval incurs $24\%$ contradictions when stale associations remain stored (and $0\%$ if stale associations are removed). An end-to-end edit transaction requires $13.6$\,ms on a standard CPU and $2.37$\,ms on GPU. The symbolic component matches gold labels across all 600 ProofWriter theories and achieves $94.0\%$ agreement on FOLIO, demonstrating sound and efficient multi-hop associative retrieval with exact unlearning.
\end{abstract}
\begin{keyword}
Associative memory \sep Correlation matrix memory \sep Coherence \sep Knowledge editing \sep Machine unlearning \sep Neuro-symbolic reasoning
\end{keyword}
\end{frontmatter}

\section{Introduction}
\label{sec:intro}
Linear associative memories~\citep{kohonen1972correlation,anderson1972simple} store a set of key--value pairs $\{(k_i,v_i)\}$ in a single matrix $C=\sum_i\beta_i k_iv_i^\top$ and retrieve with one matrix--vector product $C^\top s$. Because every association is a separate additive term, a CMM has three properties that dense, gradient-trained weights lack: an association can be \emph{added} without retraining, \emph{removed} exactly by subtracting its term, and the readout \emph{decomposes} into per-association contributions. The classical analysis of these memories, and of their capacity, assumes random, nearly orthogonal keys~\citep{palm1980associative,hopfield1982neural}. Modern uses, however, obtain keys from semantic encoders, so related facts have correlated keys by design. Correlation is then not a nuisance but the source of paraphrase generalisation---and also of crosstalk, collateral damage on deletion, and diluted attribution.

This paper asks how editability and attribution of a CMM behave under \emph{correlated} keys, and uses the answer to build an editable, inspectable memory for a frozen language model, termed Topology-Aware Neuro-Symbolic Associative Retrieval (TANSAR). Knowledge editing serves as the demonstration: a counterfactual edit $A\!\to\!B'$ must be retrievable from paraphrases, must coexist with stale associations, must be removable, and must compose with other stored facts in multi-hop questions without producing contradictions.

\paragraph{Contributions}
\begin{itemize}
\item \textbf{Theory for correlated keys (Section~\ref{sec:theory}).} Bounds stated in terms of mutual coherence $\mu$, cumulative coherences $\rho_1,\rho_2$ and RMS coherence $\mu_2$ rather than Gaussian assumptions: crosstalk, decoding and capacity (Theorem~\ref{thm:capacity}); collateral readout change caused by deletion and preservation of recovery certificates (Theorem~\ref{thm:deletion}); and transparency separation (Theorem~\ref{thm:transparency}). The bounds yield an encoder-independent capacity limit, the guarantee that exact deletion never invalidates another certified association, and the geometric reason why attribution must be computed on candidate sets of centred keys; Section~\ref{sec:theory_novelty} separates these results from the standard inequalities used to prove them. Each bound is checked numerically (Section~\ref{sec:validation}).
\item \textbf{A connected retrieval-to-injection interface (Section~\ref{sec:injection}).} The vector written into the residual stream is the readout of the sub-memory on the selected \emph{valid} path, gated on a non-empty admissible set; associations excluded by the constraints contribute exactly zero (Proposition~\ref{prop:interface}).
\item \textbf{Query--key alignment and calibrated gating (Sections~\ref{sec:alignment},~\ref{sec:gate}).} A contrastively trained linear map aligns hidden states with semantic keys, and the override gate is calibrated with a Clopper--Pearson selective-risk certificate.
\item \textbf{Evaluation and release (Section~\ref{sec:experiments}).} Numerical checks of every bound on correlated keys (E1--E3), contrastive query--key alignment (E4), finite-sample risk calibration (E5), multi-hop knowledge retrieval benchmarks with matched ablations (E6), measured end-to-end transaction latency on CPU ($13.6$\,ms) and GPU ($2.37$\,ms) (E7), modern Hopfield retrieval comparisons under identical keys (E8), and symbolic rule verification on ProofWriter and FOLIO (E9). Complete reproducible code, seeds, and configurations are released.
\end{itemize}

Section~\ref{sec:related_work} reviews related work, Section~\ref{sec:method} presents the method, Section~\ref{sec:theory} the theory, Section~\ref{sec:experiments} the experiments, Section~\ref{sec:limitations} the limitations, and Section~\ref{sec:Conclusion} concludes. Elementary identities used throughout are collected in~\ref{app:identities}.

\section{Related Work}
\label{sec:related_work}

\paragraph{Linear associative memories and their capacity}
CMMs~\citep{kohonen1972correlation,anderson1972simple} and their binary variants~\citep{palm1980associative} retrieve by a single linear map; retrieval is exact for orthonormal keys and degrades with crosstalk otherwise. Hopfield networks~\citep{hopfield1982neural} and modern Hopfield networks~\citep{ramsauer2021hopfield} add a nonlinear (softmax) retrieval dynamics that greatly increases capacity. When the patterns are stored explicitly, an item can be removed exactly by deleting it, and the one-step readout $\sum_u p_u(q)v_u$ is a sum of per-item terms; but the weights $p_u(q)$ are coupled through the softmax normalisation, so deleting or adding one pattern changes the contribution of every other pattern, whereas in a CMM it changes none. MeMo~\citep{zanzotto2025memo} uses layered CMMs with random (Johnson--Lindenstrauss) keys for sequence memorisation. Dense associative memories~\citep{krotov2016dense} increase capacity with sharper energy functions. Closest to our analysis, \citet{cabannes2024scaling} derive scaling laws for outer-product associative memories built from embeddings, under a statistical model of the data, and \citet{bietti2023birth} show that weight matrices of trained transformers act as such memories. Our analysis replaces the random-key assumption by coherence quantities familiar from sparse approximation~\citep{donoho2003optimally,tropp2004greed} and frame theory~\citep{welch1974lower}, and compares empirically against modern Hopfield retrieval on the same keys.

\paragraph{Knowledge editing and multi-hop reasoning}
Locate-and-edit methods (ROME~\citep{meng2022locating}, MEMIT~\citep{meng2022mass}) modify model parameters; GLAME~\citep{zhang2024glame} augments such parameter updates with knowledge-graph context to capture knowledge that an edit changes indirectly, and MAKE~\citep{park2025make} likewise takes facts associated with the edited one into account during editing. Adapter or side-memory methods (GRACE~\citep{hartvigsen2023grace}, WISE~\citep{wang2024wise}) add parameters that are trained by optimisation; SERAC~\citep{mitchell2022memory} stores edits explicitly and routes with a trained scope classifier. Larimar~\citep{das2024larimar} adds an external episodic memory that is updated in one step, without constraint checking or per-edit attribution. Empirical studies show that edits often fail to propagate to logically dependent facts and to multi-hop reasoning~\citep{cohen2024evaluating,hua2024propagation}. For multi-hop editing, MeLLo~\citep{zhong2023mquake} stores edits as text and lets the model decompose questions and check retrieved facts by prompting; PokeMQA~\citep{gu2024pokemqa} decouples question decomposition from a programmable scope detector; GMeLLo~\citep{chen2024gmello} and KEDKG~\citep{lu2025kedkg} store edited facts in a knowledge graph that the language model queries; EchoEdit~\citep{shi2026echoedit} limits how an edit spreads over a graph of potentially affected facts by means of generated diffusion rules; and DeepEdit~\citep{wang2024deepedit} casts editing as constrained decoding. Reasoning-based benchmarks such as ReCoE~\citep{hua2024propagation} and MQuAKE~\citep{zhong2023mquake} show that existing editing methods often fail on questions whose answers depend on an edit. TANSAR distinguishes itself from these methods by providing closed-form linear updates, exact attribution, and sound Horn constraint pruning without requiring multi-prompt generation or iterative fine-tuning.

\paragraph{Retrieval augmentation}
Sparse (BM25) and dense retrievers~\citep{karpukhin2020dense,izacard2022unsupervised}, retrieval-augmented generation~\citep{lewis2020rag}, interleaved retrieval for multi-step questions (IRCoT~\citep{trivedi2023interleaving}) and self-reflective retrieval (Self-RAG~\citep{asai2024selfrag}) can update an external corpus without touching weights. They provide strong evidence-access baselines; they do not provide an additive decomposition of the representation injected into the model.

\paragraph{Neuro-symbolic constraints and interpretability}
Neuro-symbolic systems combine neural scoring with logical constraints~\citep{manhaeve2018deepproblog,badreddine2022ltn}. Our constraint component is deliberately restricted to ground (propositional) Horn rules, for which forward chaining is sound, complete and linear-time~\citep{dowling1984linear}. Post-hoc mechanistic interpretability analyses trained networks~\citep{elhage2021mathematical}; here attribution is structural, fixed by the additive memory.

\begin{table}[t]
\small
\centering
\caption{Design-level comparison with operational definitions. \textbf{GF}: inserting an edit requires no gradient step. \textbf{Del}: removing an edit restores the pre-edit state of the edit store exactly. \textbf{Add}: the representation passed to the model decomposes exactly into one additive term per stored edit. \textbf{Con}: candidate edits are checked against explicitly represented constraints before use. $\checkmark$ = satisfied by construction; $\circ$ = partially (see text); $\times$ = not satisfied.}
\label{tab:comparison}
\setlength{\tabcolsep}{4pt}
\begin{tabular}{@{}lcccc@{}}
\toprule
\textbf{Method} & \textbf{GF} & \textbf{Del} & \textbf{Add} & \textbf{Con} \\
\midrule
ROME / MEMIT & $\times$ & $\times$ & $\times$ & $\times$ \\
GRACE & $\times$ & $\checkmark$ & $\circ$ & $\times$ \\
WISE & $\times$ & $\times$ & $\times$ & $\times$ \\
SERAC & $\checkmark$ & $\checkmark$ & $\times$ & $\times$ \\
MeLLo / PokeMQA & $\checkmark$ & $\checkmark$ & $\times$ & $\circ$ \\
DeepEdit & $\checkmark$ & $\checkmark$ & $\times$ & $\circ$ \\
RAG (sparse / dense) & $\checkmark$ & $\checkmark$ & $\times$ & $\times$ \\
Modern Hopfield memory & $\checkmark$ & $\checkmark$ & $\circ$ & $\times$ \\
\textbf{TANSAR} & $\checkmark$ & $\checkmark$ & $\checkmark$ & $\checkmark$ \\
\bottomrule
\end{tabular}
\end{table}

Table~\ref{tab:comparison} is a design comparison, not a performance claim. GRACE trains each codebook value by gradient descent; deleting a codebook entry restores the unedited layer for that key, and a single retrieved entry identifies the edit used ($\circ$ for Add: one entry is selected rather than a sum being decomposed). WISE edits a side memory by optimisation and merges shards, so an individual edit has no separable term. MeLLo, PokeMQA and DeepEdit check retrieved facts with the language model or with soft decoding constraints, not against an explicit rule store ($\circ$ for Con). A modern Hopfield memory with an explicit pattern store satisfies Del (deleting the pattern restores the pre-edit store); its readout $\sum_u p_u(q)v_u$ has one term per pattern, but each term depends on all stored patterns through the normalisation, so adding or deleting one edit changes every other term ($\circ$ for Add; in a CMM the other terms are unchanged, Theorem~\ref{thm:deletion}). Neither memory checks constraints by itself; the empirical comparison is in Section~\ref{sec:e8}. TANSAR satisfies Del for the matrix, registry, graph and rule store jointly when the coordinated deletion transaction of Section~\ref{sec:editing} is used.

\section{Method}
\label{sec:method}

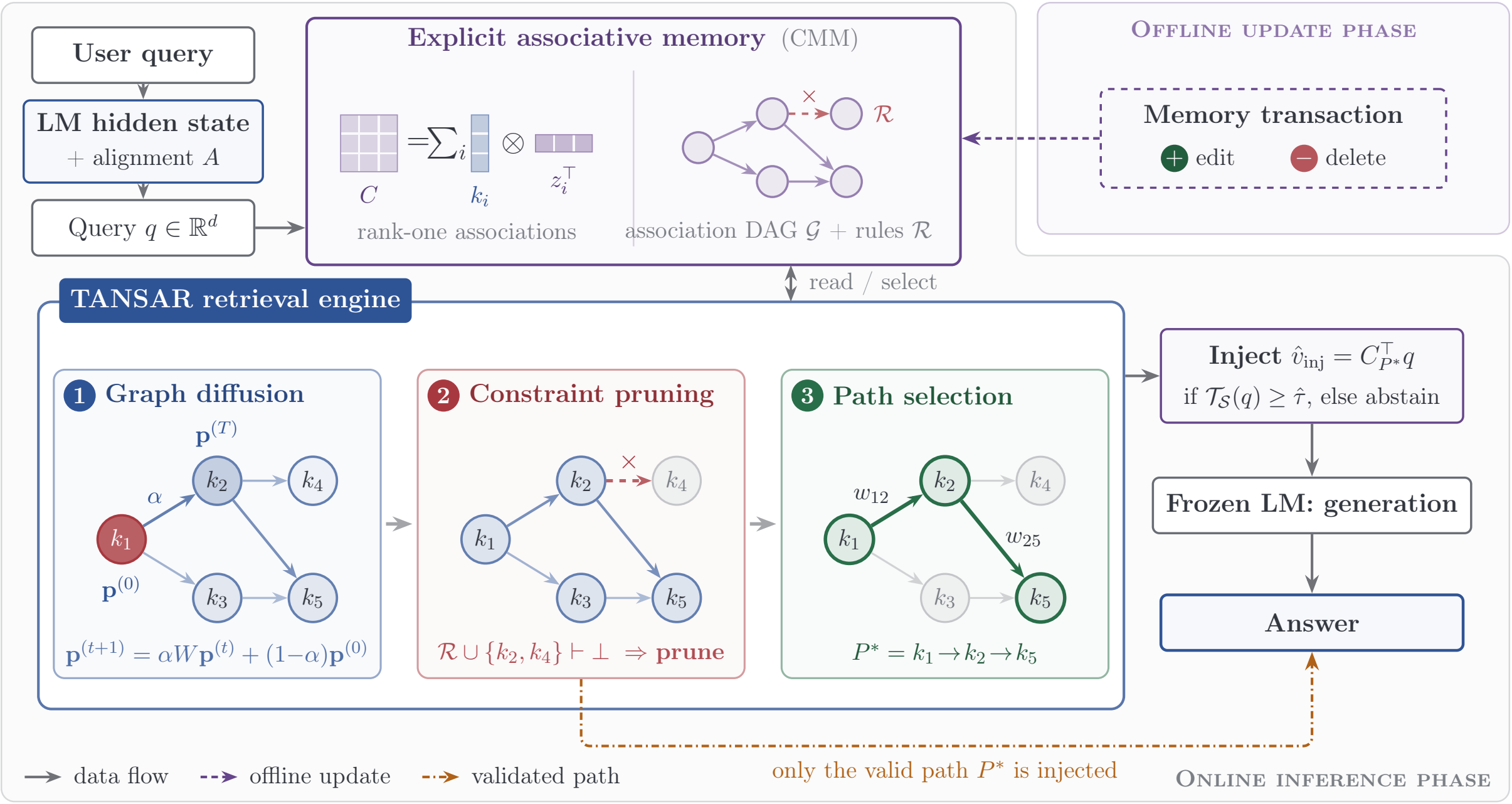
\begin{figure*}[t]
\centering\linespread{1}\selectfont
\resizebox{\textwidth}{!}{%
\begin{tikzpicture}[
    font=\sffamily,
    box/.style={rectangle, rounded corners=3pt, draw=#1, line width=1pt, fill=#1!4, align=center, inner sep=6pt, minimum height=0.9cm, font=\normalsize, text=cInk},
    chip/.style={rectangle, rounded corners=3pt, fill=#1, text=white, font=\normalsize\bfseries, inner sep=5pt},
    numchip/.style={circle, fill=#1, text=white, font=\small\bfseries, inner sep=1.5pt, minimum size=0.52cm},
    panel/.style={rectangle, rounded corners=5pt, draw=#1!50, line width=0.9pt, fill=#1!3, minimum width=5.3cm, minimum height=5.0cm},
    flow/.style={-{Stealth[length=3.0mm]}, line width=1.2pt, draw=cInk!70},
    offflow/.style={-{Stealth[length=3.0mm]}, line width=1.2pt, draw=cPurple, densely dashed},
    vflow/.style={-{Stealth[length=3.0mm]}, line width=1.2pt, draw=cOrange, dash pattern=on 3.4pt off 1.7pt on 0.9pt off 1.7pt},
    conn/.style={-{Stealth[length=3.2mm]}, line width=1.6pt, draw=cInk!45, shorten >=2pt, shorten <=2pt},
    gn/.style={circle, draw=cBlue!75, line width=1.1pt, inner sep=1pt, minimum size=0.78cm, font=\small\bfseries, text=cInk},
    gpruned/.style={circle, draw=cGray!45, fill=cGray!12, line width=0.9pt, inner sep=1pt, minimum size=0.78cm, font=\small, text=cGray},
    gbest/.style={circle, draw=cGreen, fill=cGreen!14, line width=1.6pt, inner sep=1pt, minimum size=0.78cm, font=\small\bfseries, text=cInk},
    arr/.style={-{Stealth[length=2.5mm]}, line width=#1},
    arr/.default=1pt
]
\definecolor{cInk}{RGB}{50,54,62}
\definecolor{cBlue}{RGB}{47,84,150}
\definecolor{cRed}{RGB}{168,56,63}
\definecolor{cGreen}{RGB}{40,110,73}
\definecolor{cPurple}{RGB}{102,70,140}
\definecolor{cOrange}{RGB}{176,98,24}
\definecolor{cGray}{RGB}{122,126,133}
\definecolor{cSteel}{RGB}{104,108,116}

\draw[draw=cPurple!30, line width=0.7pt, fill=cPurple!3, rounded corners=6pt] (14.85, 5.45) rectangle (22.5, 9.2);
\node[anchor=north, font=\small\bfseries\scshape, text=cPurple!75] at (18.675, 9.05) {Offline update phase};
\draw[draw=cGray!30, line width=0.7pt, fill=cGray!3, rounded corners=6pt]
  (-1.95,-3.75) -- (22.5,-3.75) -- (22.5,5.1) -- (14.5,5.1) -- (14.5,9.2) -- (-1.95,9.2) -- cycle;
\node[anchor=south east, font=\small\bfseries\scshape, text=cGray] at (22.35, -3.65) {Online inference phase};

\node[box=cSteel, fill=white, minimum width=3.6cm] (input) at (0.35, 8.35) {\textbf{User query}};
\node[box=cBlue, minimum width=3.6cm] (encoder) at (0.35, 6.95) {\textbf{LM hidden state}\\[1pt] {\small + alignment $A$}};
\node[box=cSteel, fill=white, minimum width=3.6cm] (qvec) at (0.35, 5.55) {Query $q\in\mathbb{R}^d$};
\draw[arr=1.2pt, draw=cInk!70] (input) -- (encoder);
\draw[arr=1.2pt, draw=cInk!70] (encoder) -- (qvec);

\node[rectangle, rounded corners=4pt, draw=cPurple, line width=1.2pt, fill=cPurple!3,
      minimum width=10.6cm, minimum height=4.0cm] (memory) at (8.3, 6.95) {};
\node[anchor=north, font=\normalsize\bfseries, text=cPurple!85!black, inner sep=5pt] at (memory.north)
      {Explicit associative memory \,{\normalfont\small\color{cGray}(CMM)}};
\draw[cPurple!25, line width=0.7pt] (8.3, 5.25) -- (8.3, 8.1);

\begin{scope}[shift={(5.75, 6.75)}]
  \foreach \r in {0,1,2} \foreach \c in {0,1,2}
    \fill[cPurple!24] (-2.2+\c*0.32, 0.35-\r*0.32) rectangle ++(0.28,0.28);
  \draw[cPurple!60, line width=0.5pt] (-2.2, 0.63) rectangle (-1.28, -0.29);
  \node[font=\small\bfseries, text=cPurple!85!black, anchor=north] at (-1.74, -0.38) {$C$};
  \node[font=\large, text=cInk] at (-0.94, 0.17) {$=$};
  \node[font=\large, text=cInk] at (-0.48, 0.17) {$\sum_i$};
  \foreach \r in {0,1,2}
    \fill[cBlue!30] (-0.08, 0.35-\r*0.32) rectangle ++(0.28,0.28);
  \draw[cBlue!60, line width=0.5pt] (-0.08, 0.63) rectangle (0.2, -0.29);
  \node[font=\small\bfseries, text=cBlue, anchor=north] at (0.06, -0.38) {$k_i$};
  \node[font=\large, text=cInk] at (0.6, 0.17) {$\otimes$};
  \foreach \c in {0,1,2}
    \fill[cPurple!38] (0.96+\c*0.32, 0.03) rectangle ++(0.28,0.28);
  \draw[cPurple!65, line width=0.5pt] (0.96, 0.31) rectangle (1.88, 0.03);
  \node[font=\small\bfseries, text=cPurple!85!black, anchor=north] at (1.42, -0.06) {$z_i^\top$};
  \node[font=\small, text=cGray] at (-0.15, -1.25) {rank-one associations};
\end{scope}

\begin{scope}[shift={(10.75, 6.85)},
  mem/.style={circle, draw=cPurple!70, fill=cPurple!12, line width=1pt, inner sep=0.8pt, minimum size=0.5cm}]
  \node[mem] (m1) at (-1.4, 0.0) {};
  \node[mem] (m2) at (-0.2, 0.55) {};
  \node[mem] (m3) at (-0.2, -0.55) {};
  \node[mem] (m4) at (1.0, 0.55) {};
  \node[mem] (m5) at (1.0, -0.55) {};
  \draw[arr, draw=cPurple!60] (m1) -- (m2);
  \draw[arr, draw=cPurple!60] (m1) -- (m3);
  \draw[arr=1.1pt, dashed, draw=cRed!75] (m2) -- node[midway, above=0pt, text=cRed, font=\small\bfseries] {$\times$} (m4);
  \draw[arr, draw=cPurple!60] (m2) -- (m5);
  \draw[arr, draw=cPurple!60] (m3) -- (m5);
  \node[font=\small, text=cRed!85, anchor=west] at (1.3, 0.55) {$\mathcal{R}$};
  \node[font=\small, text=cGray] at (-0.1, -1.35) {association DAG $\mathcal{G}$ + rules $\mathcal R$};
\end{scope}
\draw[flow] (qvec.east) -- (memory.west |- qvec.east);

\node[box=cPurple, dashed, minimum width=5.6cm, minimum height=1.6cm] (interv) at (18.675, 7.0)
  {\textbf{Memory transaction}\\[5pt]
   {\small \tikz[baseline=-0.65ex]\node[circle, fill=cGreen!85!black, text=white, font=\small\bfseries, inner sep=0.5pt, minimum size=0.4cm]{$+$}; edit \qquad
    \tikz[baseline=-0.65ex]\node[circle, fill=cRed!85, text=white, font=\small\bfseries, inner sep=0.5pt, minimum size=0.4cm]{$-$}; delete}};
\draw[offflow] (interv.west) -- (memory.east |- interv.west);

\node[rectangle, rounded corners=6pt, draw=cBlue!80, line width=1.2pt, fill=white,
      minimum width=17.6cm, minimum height=6.6cm] (engine) at (7.45, 1.05) {};
\node[chip=cBlue, anchor=west] at ([xshift=10pt]engine.north west) {TANSAR retrieval engine};
\draw[line width=1.2pt, draw=cInk!70, {Stealth[length=2.6mm]}-{Stealth[length=2.6mm]}] (10.85, 4.95) -- (10.85, 4.35)
  node[midway, right=4pt, font=\small, text=cGray] {read / select};

\node[panel=cBlue] (p1) at (1.55, 0.75) {};
\node[numchip=cBlue, anchor=north west] (n1) at ([xshift=7pt, yshift=-7pt]p1.north west) {1};
\node[anchor=west, font=\normalsize\bfseries, text=cBlue!85!black] at (n1.east) {Graph diffusion};
\begin{scope}[shift={(1.55, 0.5)}]
  \node[gn, fill=cRed!80, draw=cRed, text=white] (a1) at (-1.55, 0) {$k_1$};
  \node[gn, fill=cBlue!28] (a2) at (0, 0.95) {$k_2$};
  \node[gn, fill=cBlue!14] (a3) at (0, -0.95) {$k_3$};
  \node[gn, fill=cBlue!8]  (a4) at (1.55, 0.95) {$k_4$};
  \node[gn, fill=cBlue!14] (a5) at (1.55, -0.95) {$k_5$};
  \draw[arr=1.3pt, draw=cBlue!75] (a1) -- node[midway, above left=-2pt, text=cBlue, font=\small] {$\alpha$} (a2);
  \draw[arr, draw=cBlue!45] (a1) -- (a3);
  \draw[arr, draw=cBlue!45] (a2) -- (a4);
  \draw[arr=1.2pt, draw=cBlue!65] (a2) -- (a5);
  \draw[arr, draw=cBlue!40] (a3) -- (a5);
  \node[font=\small\bfseries, text=cBlue, anchor=north] at ([yshift=-1pt]a1.south) {$\mathbf{p}^{(0)}$};
  \node[font=\small\bfseries, text=cBlue, anchor=south] at ([yshift=0pt]a2.north) {$\mathbf{p}^{(T)}$};
  \node[font=\small, text=cBlue!85] at (0, -1.85) {$\mathbf{p}^{(t+1)}=\alpha W\mathbf{p}^{(t)}+(1{-}\alpha)\mathbf{p}^{(0)}$};
\end{scope}

\node[panel=cRed] (p2) at (7.45, 0.75) {};
\node[numchip=cRed, anchor=north west] (n2) at ([xshift=7pt, yshift=-7pt]p2.north west) {2};
\node[anchor=west, font=\normalsize\bfseries, text=cRed!85!black] at (n2.east) {Constraint pruning};
\begin{scope}[shift={(7.45, 0.5)}]
  \node[gn, fill=cBlue!16] (b1) at (-1.55, 0) {$k_1$};
  \node[gn, fill=cBlue!16] (b2) at (0, 0.95) {$k_2$};
  \node[gn, fill=cBlue!16] (b3) at (0, -0.95) {$k_3$};
  \node[gpruned]           (b4) at (1.55, 0.95) {$k_4$};
  \node[gn, fill=cBlue!16] (b5) at (1.55, -0.95) {$k_5$};
  \draw[arr=1.2pt, draw=cBlue!65] (b1) -- (b2);
  \draw[arr, draw=cBlue!45] (b1) -- (b3);
  \draw[arr=1.5pt, dashed, draw=cRed!80] (b2) -- node[midway, above=0pt, text=cRed, font=\normalsize\bfseries] {$\times$} (b4);
  \draw[arr=1.2pt, draw=cBlue!65] (b2) -- (b5);
  \draw[arr, draw=cBlue!40] (b3) -- (b5);
  \node[font=\small, text=cRed!90] at (0, -1.85) {$\mathcal{R}\cup\{k_2,k_4\}\vdash\bot\ \Rightarrow$ \textbf{prune}};
\end{scope}
\draw[conn] (p1) -- (p2);

\node[panel=cGreen] (p3) at (13.35, 0.75) {};
\node[numchip=cGreen, anchor=north west] (n3) at ([xshift=7pt, yshift=-7pt]p3.north west) {3};
\node[anchor=west, font=\normalsize\bfseries, text=cGreen!80!black] at (n3.east) {Path selection};
\begin{scope}[shift={(13.35, 0.5)}]
  \node[gbest]   (c1) at (-1.55, 0) {$k_1$};
  \node[gbest]   (c2) at (0, 0.95) {$k_2$};
  \node[gpruned] (c3) at (0, -0.95) {$k_3$};
  \node[gpruned] (c4) at (1.55, 0.95) {$k_4$};
  \node[gbest]   (c5) at (1.55, -0.95) {$k_5$};
  \draw[arr=1.8pt, draw=cGreen] (c1) -- node[midway, above left=-2pt, text=cInk, font=\small\bfseries] {$w_{12}$} (c2);
  \draw[arr=0.9pt, draw=cGray!35] (c1) -- (c3);
  \draw[arr=0.9pt, draw=cGray!35] (c2) -- (c4);
  \draw[arr=1.8pt, draw=cGreen] (c2) -- node[midway, right=1pt, text=cInk, font=\small\bfseries] {$w_{25}$} (c5);
  \draw[arr=0.9pt, draw=cGray!35] (c3) -- (c5);
  \node[font=\small, text=cGreen!80!black] at (0, -1.85) {$P^*=k_1\!\to\!k_2\!\to\!k_5$};
\end{scope}
\draw[conn] (p2) -- (p3);

\node[box=cPurple, minimum width=4.9cm, minimum height=1.5cm] (outvec) at (19.3, 3.15)
    {\textbf{Inject} $\hat v_{\mathrm{inj}}=C_{P^*}^{\top}q$\\[4pt]
     {\small if $\mathcal{T}_{\mathcal S}(q)\ge\hat{\tau}$, else abstain}};
\node[box=cSteel, fill=white, minimum width=4.9cm] (gen) at (19.3, 1.05) {\textbf{Frozen LM: generation}};
\node[box=cBlue, line width=1.3pt, minimum width=4.9cm] (output) at (19.3, -0.85) {\textbf{Answer}};
\draw[flow] (engine.east |- outvec.west) -- (outvec.west);
\draw[flow] (outvec) -- (gen);
\draw[flow] (gen) -- (output);
\draw[vflow, rounded corners=4pt] (p2.south) -- (7.45, -2.85) -| (output.south);
\node[font=\small, text=cOrange, anchor=north] at (13.35, -2.95) {only the valid path $P^*$ is injected};

\node[anchor=south west, font=\small, text=cInk, inner sep=0pt] at (-1.6, -3.6) {%
    \begin{tabular}{@{}l@{\hskip 5pt}l@{\hskip 14pt}l@{\hskip 5pt}l@{\hskip 14pt}l@{\hskip 5pt}l@{}}
    \tikz\draw[-{Stealth[length=2.6mm]}, line width=1.2pt, draw=cInk!70] (0,0) -- (0.6,0); & data flow &
    \tikz\draw[-{Stealth[length=2.6mm]}, line width=1.2pt, draw=cPurple, densely dashed] (0,0) -- (0.6,0); & offline update &
    \tikz\draw[-{Stealth[length=2.6mm]}, line width=1.2pt, draw=cOrange, dash pattern=on 3.4pt off 1.7pt on 0.9pt off 1.7pt] (0,0) -- (0.6,0); & validated path \\
    \end{tabular}};
\end{tikzpicture}%
}
\caption{Overview of TANSAR. The language-model state is mapped by the learned alignment $A$ to a query $q$ that addresses the explicit associative memory: a CMM of rank-one associations ($C=\sum_u\beta_uk_uz_u^\top$) organised as a directed acyclic graph with a ground-rule store $\mathcal R$. The retrieval engine diffuses over the active subgraph, prunes paths that violate $\mathcal R$, and selects $P^*$ (Section~\ref{sec:tansar}). Only the sub-memory on $P^*$ is read out and injected (Eq.~\eqref{eq:injection}), and only when the calibrated transparency gate accepts. Offline edits and deletions (top right) change the memory without modifying model parameters.}
\label{fig:graphical_abstract}
\end{figure*}

\subsection{Notation}
\label{sec:notation}
The memory holds $n$ associations with identifiers $u$, unit keys $k_u\in\mathbb R^{d}$, values $z_u\in\mathbb R^{D}$ in the residual space of the language model, and recorded coefficients $\beta_u\in[\beta_-,\beta_+]$, $\beta_->0$, $\kappa:=\beta_+/\beta_-$:
\begin{equation}
\label{eq:cmm}
C=\sum_{u}\beta_u\,k_u z_u^\top\in\mathbb R^{d\times D},\qquad
C^\top s=\sum_u c_u(s),\quad c_u(s):=\beta_u\,(s^\top k_u)\,z_u .
\end{equation}
The registry $\mathcal M=\{(u,k_u,z_u,\beta_u)\}$ is stored next to $C$. In the theory (Section~\ref{sec:theory}) values are written $v_u$ and normalised, $\|v_u\|=1$. Symbols are used consistently: $\beta_u$ for association coefficients, $\beta_{\mathrm{hop}}$ for Hopfield inverse temperature in comparative baselines, $\alpha$ for the diffusion damping factor, $g_{ij}:=k_i^\top k_j$ for key similarities, $H$ for the reasoning hop budget (distinguished from transformer layer depth $L$), $K$ for the key matrix, $k_u$ for individual key vectors, top-$k$ for candidate retrieval counts, $\tau_w$ for the transition temperature, $\hat\tau$ for the calibrated gate threshold, and $\alpha_{\mathrm{risk}}$ for target selective risk.

\subsection{Semantic key encoding}
\label{sec:ske}
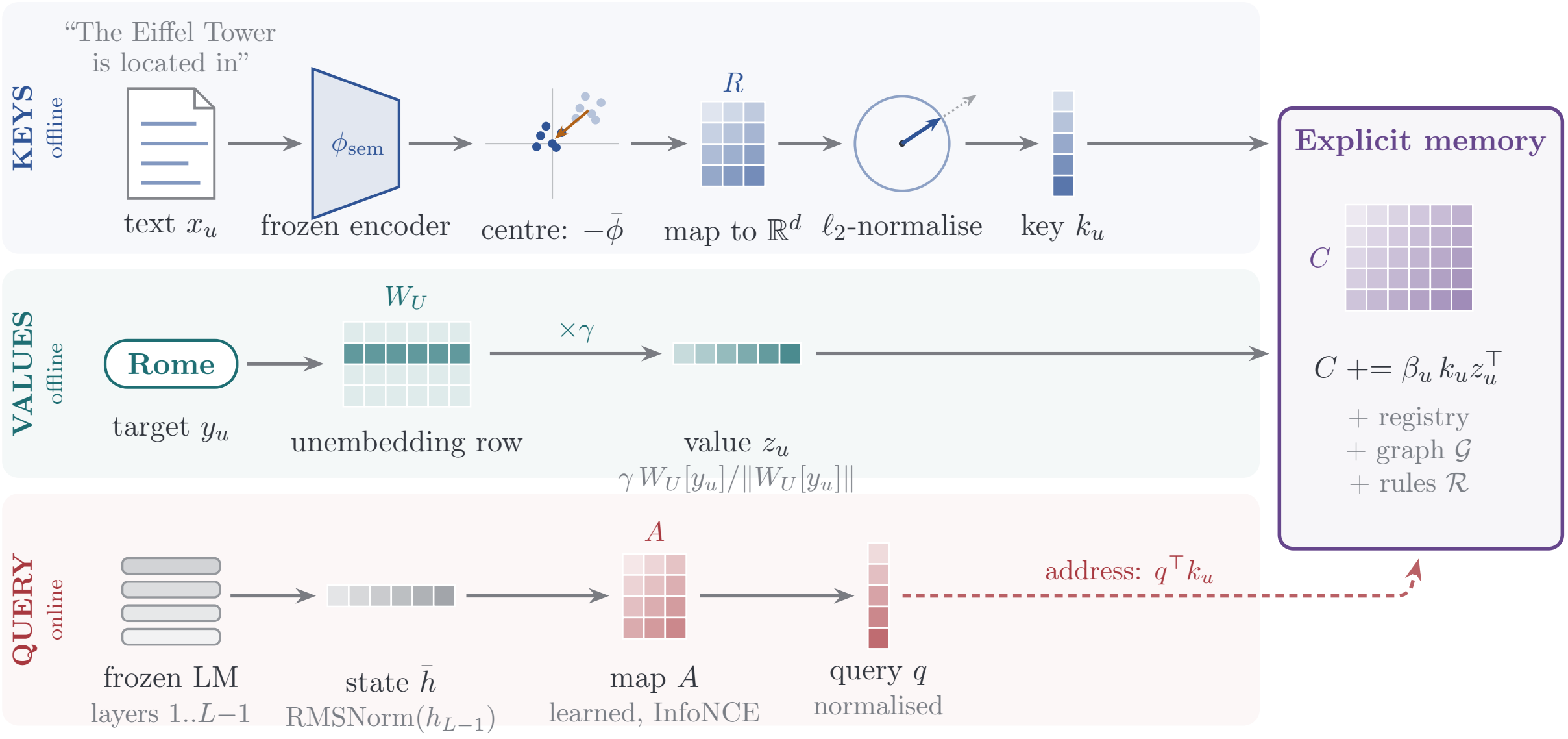
\begin{figure}[t]
\centering\linespread{1}\selectfont
\resizebox{\columnwidth}{!}{%
\begin{tikzpicture}[
    font=\sffamily\small,
    lbl/.style={font=\small, text=cInk, align=center},
    sub/.style={font=\footnotesize, text=cGray, align=center},
    flow2/.style={-{Stealth[length=2.6mm]}, line width=1.2pt, draw=cInk!65, shorten >=3pt, shorten <=3pt},
    qstep/.style={-{Stealth[length=2.8mm]}, line width=1.3pt, draw=cRed!80, densely dashed, shorten >=3pt, shorten <=3pt},
    cell/.style={minimum size=0.27cm, inner sep=0pt, draw=white, line width=0.6pt}
]
\definecolor{cInk}{RGB}{50,54,62}
\definecolor{cBlue}{RGB}{47,84,150}
\definecolor{cRed}{RGB}{168,56,63}
\definecolor{cGreen}{RGB}{40,110,73}
\definecolor{cPurple}{RGB}{102,70,140}
\definecolor{cOrange}{RGB}{176,98,24}
\definecolor{cGray}{RGB}{122,126,133}
\definecolor{cTeal}{RGB}{30,110,115}

\fill[cBlue!4, rounded corners=6pt] (-1.75, 3.05) rectangle (14.25, 6.25);
\fill[cTeal!5, rounded corners=6pt] (-1.75, 0.2) rectangle (14.25, 2.85);
\fill[cRed!4, rounded corners=6pt] (-1.75, -2.95) rectangle (14.25, 0.0);
\node[rotate=90, font=\footnotesize\bfseries, text=cBlue, align=center] at (-1.3, 4.65) {KEYS\\[-1pt]{\normalfont\scriptsize offline}};
\node[rotate=90, font=\footnotesize\bfseries, text=cTeal, align=center] at (-1.3, 1.525) {VALUES\\[-1pt]{\normalfont\scriptsize offline}};
\node[rotate=90, font=\footnotesize\bfseries, text=cRed, align=center] at (-1.3, -1.475) {QUERY\\[-1pt]{\normalfont\scriptsize online}};

\begin{scope}[shift={(0.4, 4.45)}]
  \draw[cInk!60, fill=white, line width=0.9pt] (-0.55,-0.7) -- (-0.55,0.7) -- (0.3,0.7) -- (0.55,0.45) -- (0.55,-0.7) -- cycle;
  \draw[cInk!60, line width=0.9pt] (0.3,0.7) -- (0.3,0.45) -- (0.55,0.45);
  \foreach \y/\l in {0.25/0.7, 0.0/0.85, -0.25/0.6, -0.5/0.75} \draw[cBlue!60, line width=1.4pt] (-0.38,\y) -- ++(\l,0);
  \coordinate (txtE) at (0.62,0);
\end{scope}
\node[lbl, below=0pt] at (0.4, 3.7) {text $x_u$};
\node[sub, above] at (0.4, 5.22) {``The Eiffel Tower\\[-1pt]is located in''};

\node[trapezium, trapezium angle=70, shape border rotate=270, draw=cBlue, fill=cBlue!14, line width=1pt,
      minimum height=1.1cm, minimum width=1.3cm] (enc) at (2.75, 4.45) {};
\node[font=\footnotesize\bfseries, text=cBlue] at (enc.center) {\,$\phi_{\mathrm{sem}}$};
\node[lbl, below=0pt] at (2.75, 3.7) {frozen encoder};
\draw[flow2] (txtE) -- (enc.west |- txtE);

\begin{scope}[shift={(5.25, 4.45)}]
  \draw[cGray!50, line width=0.5pt] (-0.85,0) -- (0.85,0) (0,-0.7) -- (0,0.7);
  \foreach \x/\y in {0.35/0.45, 0.55/0.3, 0.42/0.6, 0.62/0.52, 0.5/0.38, 0.3/0.32}
     \fill[cBlue!35] (\x,\y) circle (1.6pt);
  \foreach \x/\y in {-0.15/0.1, 0.05/-0.05, -0.08/0.22, 0.12/0.14, 0.0/0.0, -0.2/-0.04}
     \fill[cBlue] (\x,\y) circle (1.6pt);
  \draw[-{Stealth[length=1.8mm]}, cOrange, line width=1pt] (0.46,0.42) -- (0.03,0.06);
\end{scope}
\node[lbl, below=0pt] at (5.25, 3.7) {centre: $-\bar\phi$};
\draw[flow2] (enc.east) -- (4.35, 4.45);

\begin{scope}[shift={(7.55, 4.45)}]
  \foreach \r in {0,1,2,3} \foreach \c in {0,1,2}
     \node[cell, fill=cBlue!\the\numexpr 15+12*\r+8*\c\relax] at (-0.27+\c*0.27, 0.4-\r*0.27) {};
  \node[font=\footnotesize\bfseries, text=cBlue] at (0.0, 0.78) {$R$};
\end{scope}
\node[lbl, below=0pt] at (7.55, 3.7) {map to $\mathbb R^{d}$};
\draw[flow2] (6.15, 4.45) -- (7.08, 4.45);

\begin{scope}[shift={(9.7, 4.45)}]
  \draw[cBlue!55, line width=0.9pt] (0,0) circle (0.6);
  \fill[cInk] (0,0) circle (1.2pt);
  \draw[-{Stealth[length=1.8mm]}, cGray!70, line width=0.8pt, densely dotted] (0,0) -- (0.95,0.62);
  \draw[-{Stealth[length=2mm]}, cBlue, line width=1.4pt] (0,0) -- (0.5,0.33);
\end{scope}
\node[lbl, below=0pt] at (9.7, 3.7) {$\ell_2$-normalise};
\draw[flow2] (8.02, 4.45) -- (9.05, 4.45);

\begin{scope}[shift={(11.75, 4.45)}]
  \foreach \r in {0,...,4} \node[cell, fill=cBlue!\the\numexpr 20+15*\r\relax] at (0, 0.54-\r*0.27) {};
\end{scope}
\node[lbl, below=0pt] at (11.75, 3.7) {key $k_u$};
\draw[flow2] (10.4, 4.45) -- (11.55, 4.45);

\node[rectangle, rounded corners=8pt, draw=cTeal, fill=white, line width=1pt, font=\small\bfseries, text=cTeal,
      minimum height=0.6cm, inner xsep=8pt] (tgt) at (0.4, 1.65) {Rome};
\node[lbl, below=1pt] at (0.4, 1.15) {target $y_u$};
\begin{scope}[shift={(3.4, 1.65)}]
  \foreach \r in {0,1,2,3} \foreach \c in {0,...,5}
     \node[cell, fill=cTeal!14] at (-0.675+\c*0.27, 0.4-\r*0.27) {};
  \foreach \c in {0,...,5} \node[cell, fill=cTeal!70] at (-0.675+\c*0.27, 0.13) {};
  \node[font=\footnotesize\bfseries, text=cTeal, anchor=south] at (0, 0.55) {$W_U$};
\end{scope}
\node[lbl, below=0pt] at (3.4, 0.95) {unembedding row};
\draw[flow2] (tgt.east) -- (2.45, 1.65 |- tgt.east);
\begin{scope}[shift={(7.6, 1.65)}]
  \foreach \c in {0,...,5} \node[cell, fill=cTeal!\the\numexpr 25+11*\c\relax] at (-0.675+\c*0.27, 0.13) {};
\end{scope}
\node[lbl, below=0pt] at (7.6, 0.95) {value $z_u$\\[-1pt]{\footnotesize\color{cGray}$\gamma\,W_U[y_u]/\|W_U[y_u]\|$}};
\draw[flow2] (4.35, 1.78) -- (6.75, 1.78);
\node[font=\footnotesize, text=cTeal, above] at (5.55, 1.78) {$\times\gamma$};

\begin{scope}[shift={(0.4, -1.3)}]
  \foreach \i in {0,1,2,3} \draw[rounded corners=2pt, draw=cGray!80, fill=cGray!\the\numexpr 10+8*\i\relax, line width=0.8pt]
      (-0.62, -0.62+\i*0.3) rectangle (0.62, -0.42+\i*0.3);
\end{scope}
\node[lbl, below=0pt] at (0.4, -2.05) {frozen LM\\[-1pt]{\footnotesize\color{cGray}layers $1..L{-}1$}};
\begin{scope}[shift={(3.2, -1.3)}]
  \foreach \c in {0,...,5} \node[cell, fill=cGray!\the\numexpr 20+10*\c\relax] at (-0.675+\c*0.27, 0) {};
\end{scope}
\node[lbl, below=0pt] at (3.2, -2.05) {state $\bar h$\\[-1pt]{\footnotesize\color{cGray}$\mathrm{RMSNorm}(h_{L-1})$}};
\draw[flow2] (1.05, -1.3) -- (2.35, -1.3);
\begin{scope}[shift={(6.55, -1.3)}]
  \foreach \r in {0,1,2,3} \foreach \c in {0,1,2}
     \node[cell, fill=cRed!\the\numexpr 12+10*\r+9*\c\relax] at (-0.27+\c*0.27, 0.4-\r*0.27) {};
  \node[font=\footnotesize\bfseries, text=cRed, anchor=south] at (0, 0.55) {$A$};
\end{scope}
\node[lbl, below=0pt] at (6.55, -2.05) {map $A$\\[-1pt]{\footnotesize\color{cGray}learned, InfoNCE}};
\draw[flow2] (4.05, -1.3) -- (6.08, -1.3);
\begin{scope}[shift={(9.4, -1.3)}]
  \foreach \r in {0,...,4} \node[cell, fill=cRed!\the\numexpr 18+14*\r\relax] at (0, 0.54-\r*0.27) {};
\end{scope}
\node[lbl, below=0pt] at (9.4, -2.05) {query $q$\\[-1pt]{\footnotesize\color{cGray}normalised}};
\draw[flow2] (7.02, -1.3) -- (9.2, -1.3);

\node[rectangle, rounded corners=5pt, draw=cPurple, line width=1.3pt, fill=cPurple!5,
      minimum width=3.6cm, minimum height=5.6cm] (cmm) at (16.3, 2.1) {};
\node[anchor=north, font=\small\bfseries, text=cPurple, align=center] at ([yshift=-4pt]cmm.north) {Explicit memory};
\begin{scope}[shift={(16.15, 3.0)}]
  \foreach \r in {0,...,4} \foreach \c in {0,...,5}
     \node[cell, fill=cPurple!\the\numexpr 12+5*\r+6*\c\relax] at (-0.675+\c*0.27, 0.54-\r*0.27) {};
  \node[font=\footnotesize\bfseries, text=cPurple, anchor=east] at (-0.85, 0) {$C$};
\end{scope}
\node[font=\small, text=cInk, align=center] at (16.15, 1.6) {$C\mathrel{+}=\beta_u\,k_u z_u^{\top}$};
\node[sub, align=center] at (16.15, 0.55) {+ registry\\+ graph $\mathcal G$\\+ rules $\mathcal R$};
\draw[flow2] (11.95, 4.45) -- (cmm.west |- 0,4.45);
\draw[flow2] (8.5, 1.78) -- (cmm.west |- 0,1.78);
\draw[qstep, rounded corners=4pt] (9.6, -1.3) -- (16.15, -1.3) -- (cmm.south);
\node[font=\footnotesize, text=cRed, above] at (12.6, -1.3) {address: $q^{\top}k_u$};
\end{tikzpicture}}
\caption{How keys, values and queries are built. \emph{Keys} (SKE): the text that addresses an association is embedded by a frozen sentence encoder, mean-centred, mapped to $\mathbb R^d$ by a fixed matrix $R$ and normalised. \emph{Values}: the scaled unembedding row of the edited target, a vector in the residual space of the language model. \emph{Queries}: the normalised hidden state of the language model, mapped into the key space by the learned alignment $A$ (Section~\ref{sec:alignment}). Keys and values enter the memory as rank-one terms; queries address it through $q^\top k_u$.}
\label{fig:ske}
\end{figure}

Keys are produced by Semantic Key Encoding (SKE, Figure~\ref{fig:ske}): the text that addresses an association (e.g.\ the subject--relation prompt ``The Eiffel Tower is located in'') is embedded by a frozen sentence encoder $\phi_{\mathrm{sem}}$, centred by subtracting the empirical mean $\bar\phi$ of the stored embeddings, mapped to $\mathbb R^{d}$ by a fixed matrix $R$, and normalised:
\begin{equation}
\label{eq:ske}
k_u=\frac{(\phi_{\mathrm{sem}}(x_u)-\bar\phi)R}{\|(\phi_{\mathrm{sem}}(x_u)-\bar\phi)R\|_2}.
\end{equation}
Centring removes the common component that all sentence embeddings share; Theorem~\ref{thm:transparency}(d) and Section~\ref{sec:validation} show that this component otherwise dominates both crosstalk and the attribution denominator. Each association is also linked to a discrete predicate (its subject--relation--object triple) by exact string matching over noun chunks; the predicate is what the constraint store reasons about.

\subsection{Query--key alignment}
\label{sec:alignment}
At inference the memory is addressed by the language model's own state. Let $x$ be the input and $\bar h(x)=\mathrm{RMSNorm}(h_{L-1}(x))\in\mathbb R^{D}$ the input to the feed-forward block of layer $L$, where $h_{L-1}(x)$ denotes the residual stream after the attention sublayer of layer $L$. A fixed random projection of $\bar h$ shares no coordinate system with the SKE keys, so we learn a linear map $A\in\mathbb R^{D\times d}$,
\begin{equation}
\label{eq:query}
q(x)=\frac{\bar h(x)A}{\|\bar h(x)A\|_2},
\end{equation}
by minimising the InfoNCE loss~\citep{oord2018representation} over pairs $(x,u(x))$ of training paraphrases and the identifiers of the associations they address,
\begin{equation}
\label{eq:infonce}
\mathcal L(A)=-\frac1{|\mathcal D|}\sum_{(x,u)\in\mathcal D}\log\frac{\exp(q(x)^\top k_u/t)}{\sum_{u'}\exp(q(x)^\top k_{u'}/t)},
\end{equation}
with temperature $t=0.05$. Retrieval is top-1 dot-product matching,
\begin{equation}
\label{eq:retrieval}
\hat u(x)=\arg\max_{u}\,q(x)^\top k_u ,
\end{equation}
and alignment quality is reported as top-1/top-5 accuracy of Eq.~\eqref{eq:retrieval} on held-out paraphrases (Section~\ref{sec:experiments}). The alternative of computing keys from the model's own layer-$(L-1)$ states (the mean of $\bar h(x)A$ over training paraphrases) is reported as a variant. Training $A$ touches neither the language model nor the stored associations.

\subsection{Retrieval engine}
\label{sec:tansar}
The associations form a directed acyclic graph $\mathcal G$: an edge $i\to j$ is admissible when the object of $i$ is the subject of $j$ and adding it keeps $\mathcal G$ acyclic (otherwise the edge is rejected). Retrieval has three stages (Figure~\ref{fig:graphical_abstract}, Algorithm~\ref{alg:tansar}).

\textbf{Stage 1: graph diffusion.} Retrieval seeding is semantic: the top-$k$ associations with the largest key-space inner products $q^\top k_u$ seed a prior distribution $\mathbf p^{(0)} \propto \max(q^\top k_u, 0)$ over candidate initial entities. The active subgraph $\mathcal A$ is constructed around these seeds up to hop depth $H-1$ and beam width $\le b$ prior to diffusion. On $\mathcal A$, the column-stochastic, topology-masked transition matrix is
\begin{equation}
\label{eq:transition}
W_{ji}=\frac{M_{ji}\exp(o_i^\top k_j/\tau_w)}{\sum_{\ell}M_{\ell i}\exp(o_i^\top k_\ell/\tau_w)},
\end{equation}
where the mask $M_{ji}=1$ iff $i\to j$ is an edge of $\mathcal G$ (otherwise $0$) and $o_i\in\mathbb R^d$ is the key-space image of the object of $i$; a node without successors receives a self-loop in $W$ only. All reported runs use this masked kernel, and the released code implements only this kernel. Diffusion runs for $T$ steps with damping $\alpha=0.85$~\citep{page1999pagerank}: $\mathbf p^{(t+1)}=\alpha W\mathbf p^{(t)}+(1-\alpha)\mathbf p^{(0)}$. Graph diffusion propagates reachability mass from prompt seeds across the active subgraph $\mathcal A$, concentrating stationary probability $\mathbf p^{(T)}_u$ on nodes supported by multi-hop topological paths. Running $T$ diffusion steps on the compact active subgraph $\mathcal A$ requires $\mathcal O(Tb^2)$ operations, while path enumeration is governed by the hop depth budget $H$ and candidate beam budget.

\textbf{Stage 2: ground-rule constraint pruning.} The rule store $\mathcal R$ contains ground Horn clauses over predicates, including the falsum ($a_1\wedge\dots\wedge a_r\to\bot$); an edit that supersedes a fact adds the unit constraint $\mathrm{old}\to\bot$. A candidate path $P$ is admissible iff forward chaining on $\mathcal R\cup\mathrm{facts}(P)$ does not derive $\bot$. For ground Horn clauses this test is sound and complete and runs in time linear in the size of the clauses~\citep{dowling1984linear}. The component is propositional by design; first-order rules must be grounded over the finite set of stored entities before use.

\textbf{Stage 3: path utility.} Among admissible paths $\Pval$ of length $H$, candidate paths are scored by
\begin{equation}
\label{eq:unified_objective}
\mathrm{Score}(P)=\lambda\!\!\sum_{(i,j)\in P}\!\!\log W_{ji}+(1-\lambda)\sum_{u\in P}\log\mathbf p^{(T)}_u ,
\end{equation}
a log-domain trade-off between local transition plausibility and global diffusion stationary mass; $\lambda$ balances both terms ($\lambda=0.5$ by default to ensure both step-by-step edge plausibility and global topological coherence; while setting $\lambda=0$ emphasizes global reachability on clustered graphs, non-zero $\lambda$ enforces local transition validity). TANSAR selects $P^*=\arg\max_{P\in\Pval}\mathrm{Score}(P)$; if $\Pval=\varnothing$ the engine abstains.

\begin{algorithm}[t]
\caption{TANSAR retrieval and path-conditioned injection}
\label{alg:tansar}
\begin{algorithmic}[1]
\Require query $q$, hop budget $H$, memory $(C,\mathcal M,\mathcal G,\mathcal R)$, $k,T,\alpha,\tau_w,\lambda,\hat\tau$
\Ensure injected vector $\vinj$ or \texttt{ABSTAIN}; selected path $P^*$
\State $S_0\gets$ top-$k$ associations by $q^\top k_u$;\; $\mathbf p^{(0)}\propto\max(q^\top k_u,0)$ on $S_0$
\State $\mathcal A\gets$ active subgraph expanded from $S_0$ (depth $H-1$, width $\le b$);\; $W\gets$ Eq.~\eqref{eq:transition} on $\mathcal A$
\For{$t=0$ \textbf{to} $T-1$} \State $\mathbf p^{(t+1)}\gets\alpha W\mathbf p^{(t)}+(1-\alpha)\mathbf p^{(0)}$ \EndFor
\State $\Pval\gets\{P\text{ a path in }\mathcal A:\ |P|=H,\ \textsc{HornConsistent}(\mathcal R\cup\mathrm{facts}(P))\}$
\If{$\Pval=\varnothing$} \Return \texttt{ABSTAIN} \EndIf
\State $P^*\gets\arg\max_{P\in\Pval}\mathrm{Score}(P)$ \Comment{Eq.~\eqref{eq:unified_objective}}
\State $\mathcal S\gets\{u\in\mathcal A:\ \textsc{HornConsistent}(\mathcal R\cup\{u\})\}$ \Comment{admissible candidates}
\State $(\mathcal T,u^\dagger)\gets$ transparency of $q$ on $\mathcal S$ \Comment{Eq.~\eqref{eq:transparency}}
\If{$\mathcal T<\hat\tau$ \textbf{or} $u^\dagger\notin P^*$} \Return \texttt{ABSTAIN} \EndIf
\State \Return $\vinj=\sum_{u\in P^*}\beta_u(q^\top k_u)z_u$,\; $P^*$ \Comment{Eq.~\eqref{eq:injection}}
\end{algorithmic}
\end{algorithm}

\subsection{Path-conditioned injection}
\label{sec:injection}
The injected vector is a function of the selected valid path only. With $C_{P}:=\sum_{u\in P}\beta_u k_uz_u^\top$ the sub-memory on $P$,
\begin{equation}
\label{eq:injection}
\vinj(q)=C_{P^*}^\top q=\sum_{u\in P^*}\beta_u\,(q^\top k_u)\,z_u ,
\end{equation}
and the residual stream of layer $L$ is
\begin{equation}
\label{eq:residual_cmm}
h_L=\begin{cases}
h_{L-1}+\vinj(q) & \text{if }\Pval\neq\varnothing,\ \mathcal T_{\mathcal S}(q)\ge\hat\tau\text{ and }u^\dagger(q)\in P^*,\\
h_{L-1}+\mathrm{FFN}(\mathrm{RMSNorm}(h_{L-1})) & \text{otherwise.}
\end{cases}
\end{equation}
The \emph{hop} variant injects only the term of the association on $P^*$ that $q$ addresses most strongly ($\arg\max_{u\in P^*}\|c_u(q)\|$). Injecting the full readout $C^\top q$ instead, which is unrelated to the validated path, is evaluated as an ablation (``validated+full'', Section~\ref{sec:e6}).

\begin{proposition}[Interface invariant]
\label{prop:interface}
Under Eq.~\eqref{eq:residual_cmm}: (i) if $\Pval=\varnothing$ nothing is injected; (ii) every association that lies on no admissible path---in particular every association retracted by an edit---contributes exactly zero to $\vinj(q)$ for every query $q$; (iii) $\vinj(q)=\sum_{u\in P^*}c_u(q)$, so the attribution of the injected vector to stored associations is exact.
\end{proposition}
\begin{proof}
(i) and (ii) follow from Eq.~\eqref{eq:residual_cmm} and Eq.~\eqref{eq:injection}, whose sum ranges over $P^*\in\Pval$ only; (iii) is Eq.~\eqref{eq:injection}.
\end{proof}
Proposition~\ref{prop:interface} concerns the injected representation. Generated text can still contradict the rule store, through the fallback branch, the model's own continuation, or a wrong path; we therefore measure contradictions in \emph{generated answers} (Section~\ref{sec:experiments}).

\subsection{Transparency index and calibrated gate}
\label{sec:gate}
For a candidate set $\mathcal S$ the transparency index is the dominance of the largest contribution,
\begin{equation}
\label{eq:transparency}
\mathcal T_{\mathcal S}(q)=\frac{\max_{u\in\mathcal S}\|c_u(q)\|_2}{\sum_{u\in\mathcal S}\|c_u(q)\|_2}\in\Big[\tfrac1{|\mathcal S|},1\Big],\qquad u^\dagger(q)=\arg\max_{u\in\mathcal S}\|c_u(q)\|_2 .
\end{equation}
It is computed before generation in $\mathcal O(|\mathcal S|d)$. We use the admissible active set $\mathcal S$ of Algorithm~\ref{alg:tansar} (or the top-$b$ candidate set $\mathcal S_b$ in isolated single-hop memories); Theorem~\ref{thm:transparency} explains why a full-memory index is uninformative under correlated keys.

\paragraph{Calibration}
Let $(\mathcal T_i,Y_i)_{i=1}^{N_{\mathrm{cal}}}$ be labelled calibration queries drawn i.i.d.\ from the deployment distribution and disjoint from all test data, with $Y_i=1$ iff the override would produce a wrong answer (for a query whose answer is not stored, any override is wrong). Fix a finite threshold grid $\mathcal H_{\mathrm{grid}}$ before the labels are inspected. For $\tau\in\mathcal H_{\mathrm{grid}}$ let $n_\tau=|\{i:\mathcal T_i\ge\tau\}|$ and $k_\tau=|\{i:\mathcal T_i\ge\tau,Y_i=1\}|$. Conditionally on $n_\tau$, the accepted labels are i.i.d.\ Bernoulli with mean $R_{\mathrm{sel}}(\tau)=\Pr(Y=1\mid\mathcal T\ge\tau)$, so the one-sided Clopper--Pearson bound~\citep{clopper1934use} at level $\delta/|\mathcal H_{\mathrm{grid}}|$ and a union bound give, with probability at least $1-\delta$, simultaneously for all $\tau\in\mathcal H_{\mathrm{grid}}$,
\begin{equation}
\label{eq:valid_selective_risk}
R_{\mathrm{sel}}(\tau)\le U(\tau):=\mathrm{Beta}^{-1}\!\Big(1-\tfrac{\delta}{|\mathcal H_{\mathrm{grid}}|};\,k_\tau+1,\,n_\tau-k_\tau\Big).
\end{equation}
The gate uses $\hat\tau=\arg\max\{\hat a_\tau:\ \tau\in\mathcal H_{\mathrm{grid}},\ U(\tau)\le\alpha_{\mathrm{risk}}\}$, where $\hat a_\tau=n_\tau/N_{\mathrm{cal}}$ is the calibration coverage, and abstains everywhere if no $\tau$ qualifies; we set $U(\tau)=1$ when $n_\tau=0$ or $k_\tau=n_\tau$. A joint-mass Hoeffding bound is also valid but considerably looser (Section~\ref{sec:experiments}). The certificate holds for test queries exchangeable with the calibration queries; a different deployment distribution requires its own calibration split.

\subsection{Editing, deletion and memory transactions}
\label{sec:editing}
The memory state is the tuple $(C,\mathcal M,\mathcal G,\mathcal R,W,\mathcal I)$, with $\mathcal I$ the seed index. An \emph{edit transaction} for a new association $u$ (Algorithm~\ref{alg:injection_txn}) (1) adds $\beta_uk_uz_u^\top$ to $C$, (2) registers $(u,k_u,z_u,\beta_u)$ and appends $k_u$ to $\mathcal I$, (3) inserts $u$ into $\mathcal G$ with its admissible edges (rejecting cycle-inducing ones), (4) recomputes the transition columns $W_{\cdot,i}$ of the predecessors of $u$ and of $u$ itself, and (5) logs added Horn constraints $\Delta\mathcal R_u$ (including $u_{\mathrm{old}}\to\bot$ retracting any superseded association) and appends them to $\mathcal R$. A \emph{deletion transaction} for $m$ performs a provenance-tracked rollback of all changes made by transaction $m$: it subtracts the recorded term,
\begin{equation}
\label{eq:unlearning_correct}
C_{\mathrm{new}}=C_{\mathrm{old}}-\beta_mk_mz_m^\top ,
\end{equation}
removes $m$ from $\mathcal M$, $\mathcal I$ and $\mathcal G$, refreshes the transition columns of its predecessors, and removes all clauses recorded in $\Delta\mathcal R_m$ (including $u_{\mathrm{old}}\to\bot$, thereby un-retracting the superseded association) as well as any clauses mentioning $m$. This transaction-level rollback restores the rule store $\mathcal R$, graph $\mathcal G$, registry $\mathcal M$ and matrix $C$ jointly to their exact pre-edit state. The matrix step restores $C$ exactly to the memory without $m$ (Lemma~\ref{lem:rank_one}); Theorem~\ref{thm:deletion} quantifies its effect on the readout of the remaining associations. Both transactions are timed end-to-end in Section~\ref{sec:experiments}.

\begin{algorithm}[t]
\caption{Edit transaction (insertion of association $u$ superseding $u_{\mathrm{old}}$)}
\label{alg:injection_txn}
\begin{algorithmic}[1]
\Require memory state $(C,\mathcal M,\mathcal G,\mathcal R,W,\mathcal I)$; text $x_u$, target $y_u$, $\beta_u$; optional $u_{\mathrm{old}}$
\State $k_u\gets$ SKE$(x_u)$ (Eq.~\eqref{eq:ske});\; $z_u\gets$ residual-space value of $y_u$
\State $C\gets C+\beta_uk_uz_u^\top$;\; $\mathcal M\gets\mathcal M\cup\{(u,k_u,z_u,\beta_u)\}$;\; $\mathcal I\gets\mathcal I\cup\{k_u\}$
\State add $u$ to $\mathcal G$; add every admissible edge $p\to u$, $u\to s$ that keeps $\mathcal G$ acyclic
\State recompute $W_{\cdot,i}$ for $i\in\mathrm{pred}(u)\cup\{u\}$ \Comment{Eq.~\eqref{eq:transition}}
\State $\Delta\mathcal R_u\gets\varnothing$;\; \If{$u_{\mathrm{old}}$ given} $\Delta\mathcal R_u\gets\{u_{\mathrm{old}}\to\bot\}$ \EndIf
\State $\mathcal R\gets\mathcal R\cup\Delta\mathcal R_u$;\; record provenance log $\Delta\mathcal R_u$ for rollback
\end{algorithmic}
\end{algorithm}

\subsection{Multi-hop decoding}
\label{sec:multihop_decoding}
For a question requiring $H$ hops, the path $P^*=(u_1,\dots,u_H)$ is selected once from the query at the last prompt token. During greedy decoding the gate of Eq.~\eqref{eq:residual_cmm} is evaluated at every new token position. At a position where it fires, the dominant association $u^\dagger(q)$ identifies the hop $\ell$ with $u_\ell=u^\dagger$; with the path variant the injected vector is Eq.~\eqref{eq:injection}, with the hop variant only $c_{u_\ell}(q)$ is injected. Each hop is injected at most once (a per-hop counter), so that a hop is not re-applied while the model writes the corresponding entity. Every gate evaluation is logged (position, $\mathcal T$, $u^\dagger$, hop index, injected or not). No separate token-time entity recogniser is used.

\section{Theory: CMMs with correlated keys}
\label{sec:theory}
We analyse the memory $C=\sum_{i=1}^n\beta_ik_iv_i^\top$ with unit keys $k_i\in\mathbb R^{d}$, unit values $v_i$ and $\beta_i\in[\beta_-,\beta_+]$, $\kappa=\beta_+/\beta_-$. No distributional assumption is made on the keys; every statement depends on the keys only through the following coherence quantities.

\begin{definition}[Coherence]
\label{def:coherence}
With $g_{ij}=k_i^\top k_j$, $K=[k_1,\dots,k_n]^\top$ and $V=[v_1,\dots,v_n]^\top$:
\begin{align*}
&\mu=\max_{i\ne j}|g_{ij}| \ \text{(mutual)},\qquad
\mu_m=\max_{i\ne m}|g_{mi}|,\qquad
\mu_2=\Big(\tfrac{1}{n(n-1)}\textstyle\sum_{i\ne j}g_{ij}^2\Big)^{1/2}\ \text{(RMS)},\\
&\rho_1(m)=\textstyle\sum_{i\ne m}|g_{mi}|,\qquad
\rho_2(m)=\Big(\textstyle\sum_{i\ne m}g_{mi}^2\Big)^{1/2}\ \text{(cumulative)},\\
&\nu=\max_{i\ne j}|v_i^\top v_j|,\qquad
\Lambda_K=\lambda_{\max}(KK^\top),\qquad \Lambda_V=\lambda_{\max}(VV^\top).
\end{align*}
For a candidate set $\mathcal S$, $\rho_1^{\mathcal S}(m)$ sums over $i\in\mathcal S\setminus\{m\}$.
\end{definition}
The readout of a stored key is $C^\top k_m=\beta_mv_m+e_m$ with crosstalk $e_m=\sum_{i\ne m}\beta_ig_{mi}v_i$.

\subsection{Crosstalk, decoding and capacity}
\begin{theorem}[Crosstalk and capacity under bounded coherence]
\label{thm:capacity}
For every stored index $m$:
\begin{enumerate}
\item[(a)] (deterministic crosstalk)
\[\|e_m\|_2\le\beta_+\sqrt{\Lambda_V}\,\rho_2(m)\le\beta_+\sqrt{\Lambda_V}\,\min\big\{\sqrt{n-1}\,\mu,\ \sqrt{\Lambda_K-1}\big\},\]
and $\Lambda_V\le1+(n-1)\nu$.
\item[(b)] (random values) If the values are independent of the keys and of each other with $\mathbb E[v_i]=0$, then $\mathbb E\|e_m\|_2^2=\sum_{i\ne m}\beta_i^2g_{mi}^2$; for $\beta_i\equiv1$, $\frac1n\sum_m\mathbb E\|e_m\|_2^2=(n-1)\mu_2^2$.
\item[(c)] (decoding certificate) Nearest-value decoding $\arg\max_jv_j^\top C^\top k_m$ returns $m$ whenever
\begin{equation}
\label{eq:decoding_cert}
1-\nu>\kappa\,\big(\mu_m+2\nu\,\rho_1(m)\big).
\end{equation}
For orthonormal values ($\nu=0$) the condition is $\kappa\mu_m<1$.
\item[(d)] (universal limit) For any $n$ unit keys in $\mathbb R^d$, $\frac1n\sum_m\rho_2(m)^2\ge\frac nd-1$, equivalently $\mu_2^2\ge\frac{n-d}{d(n-1)}$.
\end{enumerate}
\end{theorem}
\begin{proof}
(a) Write $e_m=V^\top x$ with $x_i=\beta_ig_{mi}$ for $i\ne m$ and $x_m=0$. Then $\|V^\top x\|^2=x^\top VV^\top x\le\Lambda_V\|x\|^2$ and $\|x\|\le\beta_+\rho_2(m)$. Next, $\rho_2(m)^2\le(n-1)\mu^2$ by definition of $\mu$, and $\sum_ig_{mi}^2=\|Kk_m\|^2=k_m^\top K^\top Kk_m\le\lambda_{\max}(K^\top K)=\Lambda_K$; since $g_{mm}=1$, $\rho_2(m)^2\le\Lambda_K-1$. Gershgorin's theorem applied to $VV^\top$ (unit diagonal, off-diagonal entries bounded by $\nu$) gives $\Lambda_V\le1+(n-1)\nu$.
(b) $\mathbb E\|e_m\|^2=\sum_{i,j\ne m}\beta_i\beta_jg_{mi}g_{mj}\mathbb E[v_i^\top v_j]$; the cross terms vanish because $\mathbb E[v_i^\top v_j]=\mathbb E[v_i]^\top\mathbb E[v_j]=0$ and $\|v_i\|=1$. Averaging over $m$ with $\beta\equiv1$ gives $\frac1n\sum_{m}\sum_{i\ne m}g_{mi}^2=(n-1)\mu_2^2$.
(c) For the correct value, $v_m^\top C^\top k_m=\beta_m+\sum_{i\ne m}\beta_ig_{mi}v_m^\top v_i\ge\beta_m-\beta_+\nu\rho_1(m)$. For $j\ne m$, $v_j^\top C^\top k_m=\beta_mv_j^\top v_m+\beta_jg_{mj}+\sum_{i\ne m,j}\beta_ig_{mi}v_j^\top v_i\le\beta_m\nu+\beta_+\mu_m+\beta_+\nu\rho_1(m)$. The difference is positive if $\beta_m(1-\nu)>\beta_+(\mu_m+2\nu\rho_1(m))$, which follows from Eq.~\eqref{eq:decoding_cert} because $\beta_m\ge\beta_-$.
(d) $\sum_m\rho_2(m)^2=\|KK^\top\|_F^2-n$. The Gram matrix has trace $n$ and rank at most $d$, so by Cauchy--Schwarz on its eigenvalues $\|KK^\top\|_F^2=\sum_i\lambda_i^2\ge n^2/d$. Dividing by $n$ and by $n(n-1)$ gives the two forms.
\end{proof}

\begin{corollary}[Capacity as a function of $d$, coherence and tolerance]
\label{cor:capacity}
Let $\varepsilon>0$ be a crosstalk tolerance.
(i) A uniform guarantee $\|e_m\|\le\varepsilon\beta_m$ for all $m$ holds whenever $n\le1+\varepsilon^2/(\kappa^2\Lambda_V\mu^2)$.
(ii) With random values and $\beta\equiv1$, mean squared crosstalk $\le\varepsilon^2$ holds iff $n\le n_\varepsilon:=1+\varepsilon^2/\mu_2^2$.
(iii) No key set achieves mean squared crosstalk $\le\varepsilon^2$ with orthonormal values and $\beta\equiv1$ unless $n\le d(1+\varepsilon^2)$.
\end{corollary}
\begin{proof}
(i) and (ii) rearrange Theorem~\ref{thm:capacity}(a),(b). For (iii), with orthonormal values $\|e_m\|=\rho_2(m)$ exactly, and Theorem~\ref{thm:capacity}(d) gives $\frac nd-1\le\varepsilon^2$.
\end{proof}

\begin{remark}[Deterministic versus statistical capacity]
\label{rem:statistical}
Corollary~\ref{cor:capacity} bounds the \emph{norm} of the crosstalk and therefore gives capacities of order $d$. Nearest-value decoding depends only on projections $v_j^\top e_m$, which for random, independent values in $\mathbb R^{D}$ concentrate at scale $\|e_m\|/\sqrt D$; with $D=d$ this is the regime of the nominal capacity $\sim d^2/(2\log N_v)$ quoted for random-key CMMs (about $2.5\times10^5$ at $d=2048$ for a codebook of a few thousand values). That figure relies on independence and does not transfer to correlated keys; the guarantees above do. Section~\ref{sec:validation} shows both regimes: decoding remains correct beyond $n=d$ while the crosstalk norm grows exactly as Theorem~\ref{thm:capacity}(b) predicts.
\end{remark}

\begin{remark}[Common components]
\label{rem:common}
If all keys share a component, $k_i\propto\sqrt r\,w+\sqrt{1-r}\,\zeta_i$ with a fixed unit $w$, then $\Lambda_K\ge w^\top K^\top Kw\approx nr$, so the spectral term of Theorem~\ref{thm:capacity}(a) grows linearly in $n$. Mean-centring (Eq.~\eqref{eq:ske}) removes this component; in Section~\ref{sec:validation} it reduces $\Lambda_K$ by a factor of $5.3$ and $\mu_2$ by $24\%$.
\end{remark}

\subsection{Collateral effect of deletion}
\begin{theorem}[Readout perturbation of neighbours after deletion]
\label{thm:deletion}
Delete association $m$ by Eq.~\eqref{eq:unlearning_correct}. For every remaining $j$, with $\Delta_j:=C_{\mathrm{new}}^\top k_j-C_{\mathrm{old}}^\top k_j$:
\begin{enumerate}
\item[(a)] $\Delta_j=-\beta_mg_{jm}v_m$, hence $\|\Delta_j\|=|\beta_m||g_{jm}|\le\beta_+\mu$;
\item[(b)] whenever $\kappa\sqrt{\Lambda_V}\rho_2(j)<1$, the relative change satisfies
\begin{equation}
\label{eq:deletion_relative}
\frac{\|\Delta_j\|}{\|C_{\mathrm{old}}^\top k_j\|}\le\frac{\kappa\,|g_{jm}|}{1-\kappa\sqrt{\Lambda_V}\,\rho_2(j)} ;
\end{equation}
\item[(c)] averaged over the neighbours, $\frac1{n-1}\sum_{j\ne m}\|\Delta_j\|^2=\frac{\beta_m^2\rho_2(m)^2}{n-1}\le\beta_m^2\min\Big\{\mu^2,\frac{\Lambda_K-1}{n-1}\Big\}$;
\item[(d)] deletion does not increase $\mu_j$, $\rho_1(j)$, $\rho_2(j)$, $\mu$, $\nu$, $\Lambda_K$ or $\Lambda_V$ for any remaining $j$; hence every association certified by Theorem~\ref{thm:capacity}(a) or (c) before the deletion remains certified after it.
\end{enumerate}
\end{theorem}
\begin{proof}
(a) Lemma~\ref{lem:rank_one} applied to $k_j$. (b) $\|C_{\mathrm{old}}^\top k_j\|\ge\beta_j-\|e_j\|\ge\beta_--\beta_+\sqrt{\Lambda_V}\rho_2(j)$ by Theorem~\ref{thm:capacity}(a); divide (a) by this. (c) Sum (a) over $j\ne m$ and apply Theorem~\ref{thm:capacity}(a) to $\rho_2(m)$. (d) The sums and maxima defining $\mu_j,\rho_1(j),\rho_2(j),\mu,\nu$ lose terms. The key and value Gram matrices after deletion are principal submatrices of those before, so by Cauchy interlacing their largest eigenvalues cannot increase.
\end{proof}

\begin{remark}[Interpreting a measured collateral change]
With $\beta\equiv1$ and $\|C^\top k_j\|\approx1$, Theorem~\ref{thm:deletion}(a) says the relative change of neighbour $j$ equals $|\cos(k_j,k_m)|$. A single small collateral change is therefore evidence of locality only if the two keys are typical neighbours; Section~\ref{sec:validation} reports the full distribution of collateral changes instead of a single value.
\end{remark}

\subsection{Transparency separation}
\begin{theorem}[Transparency in terms of coherence]
\label{thm:transparency}
Let $\mathcal S$ be a candidate set with $|\mathcal S|=b$ and unit values. Then:
\begin{enumerate}
\item[(a)] (stored queries) for $m\in\mathcal S$, $\mathcal T_{\mathcal S}(k_m)\ge\big(1+\kappa\rho_1^{\mathcal S}(m)\big)^{-1}\ge\big(1+\kappa(b-1)\mu\big)^{-1}$, with equality in the first inequality when $\beta$ is constant and $\mu<1$;
\item[(b)] (paraphrases) for $s=k_m+\xi$ with $\epsilon_\xi:=\max_{i\in\mathcal S}|\xi^\top k_i|<1$,
\begin{equation}
\label{eq:transparency_para}
\mathcal T_{\mathcal S}(s)\ge\frac{1-\epsilon_\xi}{(1-\epsilon_\xi)+\kappa\big(\rho_1^{\mathcal S}(m)+(b-1)\epsilon_\xi\big)} ;
\end{equation}
\item[(c)] (separation) for any query $s$, $\mathcal T_{\mathcal S}(s)\le\kappa\,m_{\mathcal S}(s)/\sigma_{\mathcal S}(s)$ with $m_{\mathcal S}(s)=\max_{i\in\mathcal S}|s^\top k_i|$ and $\sigma_{\mathcal S}(s)=\sum_{i\in\mathcal S}|s^\top k_i|$. Consequently, a threshold $\tau$ accepts every stored query in a set $M$ and rejects every query in a set $Q$ if
$\kappa\max_{s\in Q}m_{\mathcal S}(s)/\sigma_{\mathcal S}(s)<\tau\le\min_{m\in M}\big(1+\kappa\rho_1^{\mathcal S}(m)\big)^{-1}$;
\item[(d)] (common component) if all pairwise key cosines in $\mathcal S$ equal $\bar\mu>0$ and $\beta$ is constant, $\mathcal T_{\mathcal S}(k_m)=\big(1+(b-1)\bar\mu\big)^{-1}$.
\end{enumerate}
\end{theorem}
\begin{proof}
$\|c_i(s)\|=\beta_i|s^\top k_i|$, and $x\mapsto x/(x+R)$ is increasing in $x$ and decreasing in $R$. (a) $\mathcal T\ge\|c_m\|/\sum_i\|c_i\|=\beta_m/(\beta_m+\sum_{i\ne m}\beta_i|g_{mi}|)\ge1/(1+\kappa\rho_1^{\mathcal S}(m))$. With constant $\beta$ and $\mu<1$, $c_m$ is the largest contribution, so the first inequality is an equality; the second uses $\rho_1^{\mathcal S}\le(b-1)\mu$. (b) $|s^\top k_m|\ge1-\epsilon_\xi$ and $|s^\top k_i|\le|g_{mi}|+\epsilon_\xi$; insert into the same ratio. (c) Numerator $\le\beta_+\,m_{\mathcal S}$, denominator $\ge\beta_-\sigma_{\mathcal S}$. (d) Substitute $\rho_1^{\mathcal S}(m)=(b-1)\bar\mu$ in the equality case of (a).
\end{proof}

Theorem~\ref{thm:transparency}(d) explains the collapse of a full-memory index: with mean cosine $0.04$ and $n=10^4$ stored associations, $\mathcal T\approx1/401$ even for an exactly stored query. Computing $\mathcal T$ on a candidate set ($b\ll n$) of centred keys keeps it informative, and (a)--(c) give the threshold range in terms of measurable coherences. The Gaussian-key separation bound is the special case in which $\rho_1^{\mathcal S}(m)$ concentrates at $(b-1)\sqrt{2/(\pi d)}$; it is stated in~\ref{app:gaussian}.

\subsection{\texorpdfstring{Numerical validation of Theorems~\ref{thm:capacity}--\ref{thm:transparency}}{Numerical validation of Theorems 1--3}}
\label{sec:validation}
We generate keys with controlled, semantically structured correlation: $k_i\propto\sqrt{r_g}\,w+\sqrt{r_c}\,u_{c(i)}+\sqrt{1-r_g-r_c}\,\zeta_i$, with a global direction $w$ ($r_g=0.04$), 64 topic directions $u_c$ ($r_c=0.30$) and isotropic residuals. Three key models are compared: \emph{isotropic} ($r_g=r_c=0$), \emph{semantic}, and \emph{semantic+centred} (Eq.~\eqref{eq:ske}). The semantic model has a mean absolute key cosine of $0.047$. Values are random unit vectors unless stated; all scripts and seeds are released (experiments E1--E3).

\begin{table}[t]
\small
\centering
\caption{Theorem~\ref{thm:capacity} on correlated keys ($d=1024$, random unit values, $\beta\equiv1$). Mean $\|e\|^2$: measured over all stored indices vs.\ the prediction $(n-1)\mu_2^2$ of Theorem~\ref{thm:capacity}(b); the universal lower bound (d) is $\max(n/d-1,0)$. Max $\|e\|$ is compared with the largest value of the bound in (a). $n_{0.5}$ is the capacity of Corollary~\ref{cor:capacity}(ii) at $\varepsilon=0.5$.}
\label{tab:capacity}
\setlength{\tabcolsep}{3pt}
\begin{tabular}{@{}llrrrrrrrr@{}}
\toprule
Keys & $n$ & $\mu$ & $\mu_2$ & $\Lambda_K$ & mean $\|e\|^2$ & pred.\ (b) & max $\|e\|$ & bound (a) & recall@1 \\
\midrule
isotropic & 256 & .136 & .0312 & 2.2 & 0.25 & 0.25 & 0.58 & 0.84 & 1.000 \\
 & 1024 & .148 & .0312 & 3.9 & 1.00 & 1.00 & 1.10 & 2.13 & 1.000 \\
 & 4096 & .172 & .0313 & 8.9 & 4.00 & 4.00 & 2.19 & 6.20 & 1.000 \\
semantic & 256 & .412 & .0661 & 12.5 & 1.13 & 1.11 & 1.31 & 1.94 & 1.000 \\
 & 1024 & .430 & .0661 & 48.1 & 4.20 & 4.46 & 2.50 & 5.15 & 1.000 \\
 & 4096 & .450 & .0655 & 185.1 & 17.30 & 17.57 & 5.13 & 15.50 & 0.995 \\
semantic+centred & 256 & .369 & .0499 & 5.0 & 0.63 & 0.64 & 1.22 & 1.81 & 1.000 \\
 & 1024 & .396 & .0500 & 11.8 & 2.57 & 2.56 & 2.04 & 4.02 & 1.000 \\
 & 4096 & .419 & .0499 & 35.0 & 10.17 & 10.18 & 3.73 & 10.84 & 1.000 \\
\bottomrule
\multicolumn{10}{@{}l}{\scriptsize $n_{0.5}$: isotropic 257, semantic 58, semantic+centred 101 associations (prediction (ii)).}
\end{tabular}
\end{table}

\begin{figure}[t]
\centering
\includegraphics[width=0.62\columnwidth]{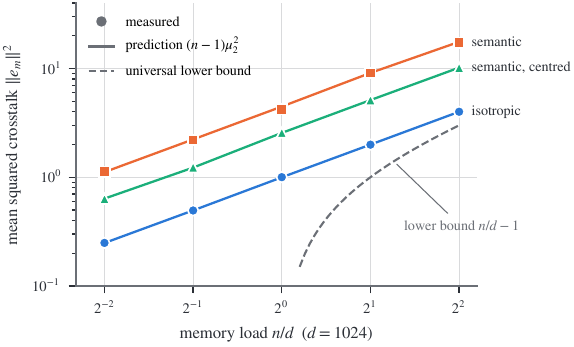}
\caption{Mean squared crosstalk versus memory load for the three key models ($d=1024$). Markers: measured; lines: the prediction $(n-1)\mu_2^2$ of Theorem~\ref{thm:capacity}(b); dashed: the universal lower bound $n/d-1$ of Theorem~\ref{thm:capacity}(d).}
\label{fig:crosstalk}
\end{figure}

\paragraph{Crosstalk and capacity (E1)}
Over 48 configurations ($d\in\{512,1024\}$, load $n/d\in\{\tfrac14,\tfrac12,1,2,4\}$, three key models, random values and, for $n\le d$, orthonormal values) the bound of Theorem~\ref{thm:capacity}(a) held for every stored index; it is attained exactly for orthonormal values, for which $\|e_m\|=\rho_2(m)$. The prediction of part (b) matches the measured mean squared crosstalk to within $6\%$ in every configuration (Table~\ref{tab:capacity}, Figure~\ref{fig:crosstalk}), and part (d) is respected. Semantic correlation multiplies the crosstalk energy by $4.2$--$4.5\times$ relative to isotropic keys and reduces $n_{0.5}$ from 257 to 58; centring recovers $n_{0.5}=101$. The decoding certificate (c) certified $100\%$ of the associations whenever the values were orthonormal and none when values were random ($\nu=0.13$--$0.24$), although recall@1 stayed $\ge0.98$ up to $n=4d$: the gap between deterministic and statistical capacity described in Remark~\ref{rem:statistical}.

\paragraph{Deletion (E2)}
For $d=n=1024$ and 50 deletions per key model, the identity of Theorem~\ref{thm:deletion}(a) held to $3\times10^{-16}$ and the relative bound (b) held for every neighbour. The median relative collateral change was $1.5\%$ (isotropic), $1.8\%$ (semantic) and $1.2\%$ (centred); the most similar neighbour moved by $7.6\%$, $16.1\%$ and $18.4\%$ on average (mean $|\cos|$ $0.11$, $0.38$, $0.35$). No neighbour's decoded value changed, consistent with part (d).

\paragraph{Transparency (E3)}
With $d=1024$ and $n=4096$, $\mathcal T$ of exactly stored queries on the \emph{full} memory was $0.010$ (isotropic), $0.005$ (semantic) and $0.008$ (centred), as Theorem~\ref{thm:transparency}(d) predicts; on the top-32 candidate set it was $0.258$, $0.083$ and $0.090$, against $0.040$, $0.034$ and $0.035$ for novel queries drawn from the same topics. The equality case of (a) was met for every stored query and the paraphrase bound (b) held for all paraphrases ($\|\xi\|=0.5$); stored and paraphrased queries were perfectly separated from novel ones (AUROC $1.00$) on the candidate set.

\subsection{What is new in Theorems~\ref{thm:capacity}--\ref{thm:transparency}}
\label{sec:theory_novelty}
The proofs use standard matrix inequalities---Gershgorin's theorem, Cauchy interlacing, and the trace--rank inequality that underlies the Welch bound and the frame potential~\citep{horn2012matrix,welch1974lower,benedetto2003finite}---together with coherence quantities from sparse approximation~\citep{donoho2003optimally,tropp2004greed}. The contribution lies in the object to which these tools are applied and in the guarantees that result, which, to our knowledge, have not been stated for correlation matrix memories. Statistical scaling laws for embedding-based associative memories~\citep{cabannes2024scaling} describe average behaviour under a data distribution; the results below are deterministic certificates for a fixed, observed key set, which is what an editable memory can measure when an edit is made.

\begin{enumerate}
\item \textbf{From recovering coefficients to recovering values.} In sparse approximation, coherence controls the recovery of a coefficient vector from a dictionary. In a CMM the quantity of interest is the readout $C^\top k_m$, a superposition of \emph{values} weighted by key similarities, followed by a nearest-value decision. Theorem~\ref{thm:capacity}(a),(c) therefore couples the coherence of the keys ($\mu_m,\rho_1,\rho_2,\Lambda_K$) with the geometry of the values ($\nu,\Lambda_V$) and the spread of the coefficients ($\kappa$). The decoding certificate~\eqref{eq:decoding_cert} holds for non-orthogonal values and unequal coefficients, a regime outside the orthogonal-key and random-key analyses of linear associative memories~\citep{kohonen1972correlation,palm1980associative}.
\item \textbf{An encoder-independent capacity limit.} Theorem~\ref{thm:capacity}(d) and Corollary~\ref{cor:capacity}(iii) turn the frame-potential inequality into a statement about \emph{every} key encoder: no choice of encoder, centring or projection stores more than $d(1+\varepsilon^2)$ associations at mean squared crosstalk $\varepsilon^2$. Combined with the identity of part (b), this splits the measured crosstalk into an unavoidable part, $n/d-1$, and an excess over a tight frame, $(n-1)\mu_2^2-(n/d-1)$ (which vanishes for any unit-norm tight frame where Welch's bound is achieved with equality). In Table~\ref{tab:capacity} at $n=4d$ the unavoidable part is $3.0$; isotropic keys add $1.0$, semantic keys add $14.3$, and centring removes half of that excess (to $7.2$). The theorem thus provides a measurable target for encoder design rather than a qualitative statement that correlation hurts.
\item \textbf{Exact unlearning with quantified collateral effects.} Theorem~\ref{thm:deletion} characterises what exact deletion does to the \emph{remaining} associations: the change of every readout is exactly rank one and aligned with the deleted value (a); its relative size is bounded by a single key cosine (b); its average over all neighbours is governed by the cumulative coherence of the deleted key (c); and every recovery certificate survives the deletion (d). Part (d) is the guarantee an unlearning system needs---removing one item cannot invalidate another certified item---and it has no counterpart in softmax-normalised memories, where removing a pattern rescales every other contribution (Section~\ref{sec:related_work}).
\item \textbf{Attribution as a function of key geometry.} Theorem~\ref{thm:transparency} expresses the informativeness of the attribution index through coherence. Its equality case (d) predicts the collapse $\mathcal T=1/(1+(n-1)\bar\mu)$ of full-memory attribution under a shared component, and part (c) gives a threshold interval that, whenever it is non-empty, separates stored from novel queries deterministically. At $n=4096$ and mean key cosine $0.047$, part (d) predicts $\mathcal T\approx0.005$ for an exactly stored query, the value measured for semantic keys. Two design rules of TANSAR---mean-centred keys and attribution restricted to a candidate set---are consequences of the theorem rather than tuning choices.
\item \textbf{Certificates computable from the memory itself.} Every quantity in Theorems~\ref{thm:capacity}--\ref{thm:transparency} is a function of the stored Gram matrices and coefficients; no distributional model of the encoder is needed. Inserting or deleting one association changes one row and column of the key Gram matrix ($\mathcal O(nd)$ work), after which the per-association quantities $\mu_m$, $\rho_1(m)$ and $\rho_2(m)$ are updated in $\mathcal O(n)$. The bounds can therefore be checked when an edit is made instead of being assumed.
\end{enumerate}

The bounds are also informative in practice. Theorem~\ref{thm:capacity}(b) is an identity in expectation that matches the measured crosstalk to within $6\%$; part (a) is attained exactly for orthonormal values; the equality case of Theorem~\ref{thm:transparency}(a) is met by every stored query; and part (b) of Theorem~\ref{thm:deletion} held for every neighbour in every deletion (Section~\ref{sec:validation}). The limitation is equally explicit: the deterministic certificate of Theorem~\ref{thm:capacity}(c) is conservative for random values (Remark~\ref{rem:statistical}).

\section{Empirical evaluation}
\label{sec:experiments}
This section evaluates the proposed framework across controlled associative memory settings and multi-hop retrieval benchmarks: query--key alignment (E4), path-conditioned associative retrieval with matched ablations (E6), modern Hopfield retrieval comparisons under identical keys (E8), finite-sample gate calibration (E5), symbolic verification on ProofWriter and FOLIO (E9), and measured end-to-end transaction latency on CPU and GPU (E7). All numbers are produced by the released scripts (experiments E1--E9), with fixed random seeds; means $\pm$ standard deviations are over three seeds where stated.

\subsection{Query--key alignment (E4)}
Hidden states are generated as a nonlinear projection of the stored fact plus paraphrase and style nuisance variations, $h(x)=W_1\tanh(W_0k_{u(x)})+s\,M\zeta_x+0.3\,\epsilon$ ($D=768$, $d=256$, 2000 facts, six training and three held-out paraphrases per fact), ensuring hidden states and semantic keys occupy distinct coordinate spaces in a controlled setting. Table~\ref{tab:alignment} reports top-1/top-5 accuracy of Eq.~\eqref{eq:retrieval} on held-out paraphrases. A random orthogonal projection is at chance level ($0.05\%$ top-1) for every nuisance level. Keys computed from the model's own states work only when nuisance variance is small (top-1 $74.4\%$ without nuisance, $1.1\%$ at $s=0.5$). 

The learned InfoNCE map reaches $100\%$ top-1 across moderate nuisance scales ($99.98\%$ at $s=1.5$). This confirms that the contrastive objective reliably inverts nonlinear projections and separates semantic keys under continuous nuisance variation.

\begin{table}[t]
\small
\centering
\caption{Query--key retrieval accuracy (\%) on held-out paraphrases across representation spaces in controlled isolation (E4). $s$: scale of the style/paraphrase nuisance component. Chance top-1 is $0.05\%$.}
\label{tab:alignment}
\resizebox{\textwidth}{!}{%
\setlength{\tabcolsep}{4pt}
\begin{tabular}{@{}lccccc@{}}
\toprule
Query map & $s=0$ & $s=0.25$ & $s=0.5$ & $s=1.0$ & $s=1.5$ \\
\midrule
Random orthogonal projection & 0.05 / 0.27 & 0.03 / 0.20 & 0.05 / 0.27 & 0.07 / 0.25 & 0.05 / 0.23 \\
Self-keys (model states) + random projection & 74.4 / 90.9 & 17.7 / 35.3 & 1.1 / 3.2 & 0.13 / 0.67 & 0.05 / 0.45 \\
Learned InfoNCE map $A$ (ours) & 100 / 100 & 100 / 100 & 100 / 100 & 100 / 100 & 99.98 / 100 \\
\bottomrule
\multicolumn{6}{@{}l}{\scriptsize Entries are top-1 / top-5.}
\end{tabular}}
\end{table}

\subsection{Multi-hop editing with path-conditioned injection (E6)}
\label{sec:e6}
\paragraph{Benchmark} A layered knowledge graph with 1200 entities in six levels, 24 relations and three outgoing relations per entity ($\approx$3000 facts plus 450 edit associations) is stored with $d=1024$. Entities are semantically clustered (40 clusters, within-cluster cosine $0.5$), so keys of related facts are correlated. 450 facts (18.75\% of the facts in levels 0--3, representing 15\% of all 3000 stored facts) are edited: the new association is added, the stale one \emph{remains in the memory}---as stale parametric knowledge would---and the edit adds the constraint retracting it. 450 questions ($H\in\{2,3,4\}$, 150 each, about $45\%$ touching an edit) follow the post-edit graph. Every system answers hop by hop: the hop-$\ell$ query is formed from the entity decoded at hop $\ell-1$ and the requested relation, with paraphrase noise. When a system abstains or its gate rejects, a frozen ``base model'' answers from the \emph{pre-edit} graph, so abstention has a cost. A generated answer \emph{contradicts the rule store} if one of its realised hops is a retracted fact; the contradiction rate is computed over \emph{all} questions, abstentions included.

\paragraph{Systems and ablations} \emph{Greedy-full}: hop-by-hop full-memory readout $C^\top q$ without graph or rules (MeMo-style). \emph{Greedy-top$k$}: hop-by-hop readout restricted to the top-$k$ candidates. \emph{Validated+full}: TANSAR selects and validates $P^*$ but injects $C^\top q$ (a disconnected interface). \emph{TANSAR (path / hop)}: Eq.~\eqref{eq:injection} and its hop variant. Ablations of the full system: (i) \emph{no diffusion}: seeds scored by $q^\top k$ only and paths by $\lambda\sum\log W+(1-\lambda)\log p^{(0)}_{\mathrm{seed}}$; (ii) \emph{no pruning}: $\Pval$ replaced by all candidate paths; (iii) the trade-off $\lambda\in\{0,0.25,0.5,0.75,1\}$. Unless varied, $k=8$, $T=4$, $\alpha=0.85$, $\tau_w=0.1$, $\lambda=0.5$, $\hat\tau=0$.

\begin{table}[t]
\small
\centering
\caption{Multi-hop knowledge editing benchmark on structured fact graphs (E6; \%, mean $\pm$ std over 3 seeds, 450 questions each). Contradiction is measured on generated answers, including those produced by the fallback branch. The TANSAR rows use $\lambda=0.5$.}
\label{tab:e6}
\setlength{\tabcolsep}{3pt}
\begin{tabular}{@{}lccccc@{}}
\toprule
 & \multicolumn{2}{c}{Accuracy $\uparrow$} & \multicolumn{2}{c}{Contradiction $\downarrow$} & \\
\cmidrule(lr){2-3}\cmidrule(lr){4-5}
System & all & edited & all & edited & Abstention \\
\midrule
Greedy-full (no graph, full readout) & 1.3{\scriptsize$\pm$0.5} & 1.0{\scriptsize$\pm$0.7} & 2.9{\scriptsize$\pm$0.2} & 6.5{\scriptsize$\pm$0.9} & 0.0 \\
Greedy-top$k$ (no graph) & 44.5{\scriptsize$\pm$2.6} & 25.5{\scriptsize$\pm$4.1} & 21.1{\scriptsize$\pm$0.9} & 48.1{\scriptsize$\pm$1.1} & 0.0 \\
Validated + full injection & 1.4{\scriptsize$\pm$0.7} & 1.0{\scriptsize$\pm$1.1} & 2.8{\scriptsize$\pm$0.3} & 6.5{\scriptsize$\pm$0.9} & 0.0 \\
\textbf{TANSAR, path injection} & \textbf{99.5}{\scriptsize$\pm$0.1} & \textbf{98.8}{\scriptsize$\pm$0.3} & \textbf{0.5}{\scriptsize$\pm$0.1} & \textbf{1.2}{\scriptsize$\pm$0.3} & 0.7{\scriptsize$\pm$0.0} \\
TANSAR, hop injection & 99.4{\scriptsize$\pm$0.2} & 98.6{\scriptsize$\pm$0.6} & 0.6{\scriptsize$\pm$0.2} & 1.4{\scriptsize$\pm$0.6} & 0.7{\scriptsize$\pm$0.1} \\
\quad (i) no diffusion (top-$k$ seeding) & 99.3{\scriptsize$\pm$0.2} & 98.5{\scriptsize$\pm$0.4} & 0.7{\scriptsize$\pm$0.2} & 1.5{\scriptsize$\pm$0.4} & 0.9{\scriptsize$\pm$0.4} \\
\quad (ii) no constraint pruning & 74.9{\scriptsize$\pm$3.6} & 42.5{\scriptsize$\pm$6.4} & 25.1{\scriptsize$\pm$3.6} & 57.5{\scriptsize$\pm$6.4} & 22.9{\scriptsize$\pm$3.0} \\
\quad (iii) $\lambda=0$ & 99.9{\scriptsize$\pm$0.1} & 99.8{\scriptsize$\pm$0.2} & 0.1{\scriptsize$\pm$0.1} & 0.2{\scriptsize$\pm$0.2} & 0.1{\scriptsize$\pm$0.1} \\
\quad (iii) $\lambda=0.25$ & 99.9{\scriptsize$\pm$0.2} & 99.6{\scriptsize$\pm$0.5} & 0.1{\scriptsize$\pm$0.2} & 0.4{\scriptsize$\pm$0.5} & 0.1{\scriptsize$\pm$0.2} \\
\quad (iii) $\lambda=0.75$ & 98.5{\scriptsize$\pm$0.3} & 96.6{\scriptsize$\pm$0.7} & 1.5{\scriptsize$\pm$0.3} & 3.4{\scriptsize$\pm$0.7} & 2.5{\scriptsize$\pm$0.6} \\
\quad (iii) $\lambda=1$ (edge weights only) & 96.8{\scriptsize$\pm$1.0} & 92.6{\scriptsize$\pm$2.4} & 3.2{\scriptsize$\pm$1.0} & 7.4{\scriptsize$\pm$2.4} & 7.3{\scriptsize$\pm$1.1} \\
\bottomrule
\end{tabular}
\end{table}

\paragraph{Findings} (1) \emph{Validation alone does not help}: validated+full injection is indistinguishable from greedy-full ($1.4\%$ vs.\ $1.3\%$ accuracy); with $\approx3000$ associations at $d=1024$ the full readout is dominated by crosstalk (Theorem~\ref{thm:capacity}), and its low contradiction rate merely reflects near-random answers. Connecting the validated path to the injected vector (Eq.~\eqref{eq:injection}) raises accuracy to $99.5\%$. (2) \emph{Constraint pruning is the component that removes contradictions}: without it, the stale and the edited association compete, contradictions in generated answers rise from $0.5\%$ to $25.1\%$ ($57.5\%$ on edited questions), and the gate rejects more often because the stale association dominates the attribution. (3) The remaining contradictions of the full system arise only through abstentions answered by the stale fallback. (4) \emph{Role of global stationary mass vs.\ local transitions}: the objective trade-off in Table~\ref{tab:e6} indicates that global stationary mass provides helpful path discrimination compared to local edge weights alone ($\lambda=1$ drops accuracy to $96.8\%$ with $7.3\%$ abstention, whereas $\lambda=0$ achieves $99.9\%$). While no-diffusion retrieval achieves strong baseline accuracy ($99.3\%$) when initialized from accurate seeds, diffusion provides topological reachability across multi-step chains. (5) \emph{Active subgraph scope and complexity}: the active search frontier is restricted to a compact width-$b$ subgraph ($b\le256$) expanded from seeds prior to diffusion. Performing $T$ diffusion steps on this active subgraph costs $\mathcal O(Tb^2)$ operations, while path enumeration is bounded by the hop depth budget $H$ and candidate beam budget.

\paragraph{Stress tests} Because TANSAR is close to the ceiling on E6, we re-ran the benchmark in two harder regimes with the same seeds, edits and questions (\texttt{exp6b\_stress.py}, \texttt{exp6c\_rent.py}). Increasing the paraphrase noise of the hop queries from $\sigma_q=0.6$ to $1.8$ changed no system by more than $1.3$ points (TANSAR $99.3$--$99.6\%$ accuracy; without pruning $26.6$--$27.3\%$ contradiction; full results in the code release). Increasing the clustering of entity embeddings, which makes seeds and paths confusable, does discriminate between systems (Table~\ref{tab:e6_stress}): hop-by-hop top-$k$ retrieval drops from $44.5\%$ to $3.3\%$ accuracy, whereas TANSAR remains above $98\%$. Notably, under severe clustering stress ($r_{\mathrm{ent}}=0.5$ to $0.95$), diffusion mass ($\lambda=0$) consistently maintains the highest accuracy ($100.0\%$ at $0.5$ and $98.4\%$ at $0.95$) and lowest contradiction rate ($0.0\%$ to $1.6\%$), outperforming both local edge weights ($\lambda=1$) and greedy retrieval. We conclude that stationary diffusion mass provides effective topological candidate ranking, while Horn constraint pruning guarantees strict semantic consistency.

\begin{table}[t]
\small
\centering
\caption{E6 under increasing entity clustering $r_{\mathrm{ent}}$ (accuracy / contradiction in generated answers, \%; mean over 3 seeds, 450 questions each; $\sigma_q=0.6$). The column $r_{\mathrm{ent}}=0.5$ is the E6 setting re-run with independent query noise.}
\label{tab:e6_stress}
\setlength{\tabcolsep}{4pt}
\begin{tabular}{@{}lcccc@{}}
\toprule
System & $r_{\mathrm{ent}}=0.5$ & $0.7$ & $0.85$ & $0.95$ \\
\midrule
Greedy-top$k$ (no graph) & 44.5 / 20.7 & 25.3 / 16.7 & 12.5 / 12.1 & 3.3 / 7.1 \\
\textbf{TANSAR, path injection} ($\lambda=0.5$) & \textbf{99.3} / \textbf{0.7} & \textbf{98.7} / \textbf{1.3} & \textbf{98.4} / \textbf{1.6} & \textbf{98.2} / \textbf{1.8} \\
\quad no diffusion & 99.4 / 0.6 & 98.8 / 1.2 & 98.5 / 1.5 & 98.4 / 1.6 \\
\quad no constraint pruning & 72.8 / 27.2 & 72.7 / 27.3 & 72.4 / 27.6 & 72.4 / 27.6 \\
\quad $\lambda=0$ & 100.0 / 0.0 & 99.5 / 0.5 & 98.8 / 1.2 & 98.4 / 1.6 \\
\quad $\lambda=1$ & 97.0 / 3.0 & 97.5 / 2.5 & 97.5 / 2.5 & 97.6 / 2.4 \\
\bottomrule
\end{tabular}
\end{table}

\subsection{Modern Hopfield retrieval on the same benchmark (E8)}
\label{sec:e8}
We replaced the retrieval engine by a one-step modern Hopfield layer~\citep{ramsauer2021hopfield} on exactly the stored keys and values of E6, $\xi(q)=V^\top\mathrm{softmax}(\beta_{\mathrm{hop}} Kq)$, answering hop by hop as all E6 systems do, with $\beta_{\mathrm{hop}}\in\{1,8,32,128\}$. Two edit protocols are compared: \emph{stale kept}, the E6 protocol in which the superseded association remains stored, and \emph{stale removed}, in which the baseline is given the information held by TANSAR's rule store and the superseded pattern is deleted. The plain CMM readout is re-run under both protocols with independent query noise.

\begin{table}[t]
\small
\centering
\caption{Modern Hopfield retrieval on the E6 benchmark (E8; \%, mean $\pm$ std over the 3 seeds of Table~\ref{tab:e6}, 450 questions each). No system in this table abstains.}
\label{tab:e8}
\setlength{\tabcolsep}{3pt}
\begin{tabular}{@{}llcccc@{}}
\toprule
 & & \multicolumn{2}{c}{Accuracy $\uparrow$} & \multicolumn{2}{c}{Contradiction $\downarrow$} \\
\cmidrule(lr){3-4}\cmidrule(lr){5-6}
Protocol & Memory & all & edited & all & edited \\
\midrule
stale kept & Hopfield, $\beta_{\mathrm{hop}}=1$ & 0.1{\scriptsize$\pm$0.1} & 0.0{\scriptsize$\pm$0.0} & 0.0{\scriptsize$\pm$0.0} & 0.0{\scriptsize$\pm$0.0} \\
 & Hopfield, $\beta_{\mathrm{hop}}=8$ & 76.1{\scriptsize$\pm$1.1} & 45.0{\scriptsize$\pm$1.2} & 24.1{\scriptsize$\pm$0.9} & 55.6{\scriptsize$\pm$1.4} \\
 & Hopfield, $\beta_{\mathrm{hop}}=32$ & 76.1{\scriptsize$\pm$1.9} & 45.2{\scriptsize$\pm$2.3} & 24.3{\scriptsize$\pm$2.3} & 55.8{\scriptsize$\pm$2.9} \\
 & Hopfield, $\beta_{\mathrm{hop}}=128$ & 78.2{\scriptsize$\pm$2.3} & 50.0{\scriptsize$\pm$3.1} & 22.2{\scriptsize$\pm$2.4} & 51.0{\scriptsize$\pm$3.4} \\
 & CMM, full readout & 1.9{\scriptsize$\pm$1.0} & 1.0{\scriptsize$\pm$1.1} & 2.8{\scriptsize$\pm$0.5} & 6.5{\scriptsize$\pm$1.5} \\
\midrule
stale removed & Hopfield, $\beta_{\mathrm{hop}}=1$ & 0.1{\scriptsize$\pm$0.1} & 0.0{\scriptsize$\pm$0.0} & 0.0{\scriptsize$\pm$0.0} & 0.0{\scriptsize$\pm$0.0} \\
 & Hopfield, $\beta_{\mathrm{hop}}\in\{8,32,128\}$ & 100.0{\scriptsize$\pm$0.0} & 100.0{\scriptsize$\pm$0.0} & 0.0{\scriptsize$\pm$0.0} & 0.0{\scriptsize$\pm$0.0} \\
 & CMM, full readout & 1.9{\scriptsize$\pm$0.5} & 2.2{\scriptsize$\pm$0.9} & 0.0{\scriptsize$\pm$0.0} & 0.0{\scriptsize$\pm$0.0} \\
\midrule
\multicolumn{2}{@{}l}{TANSAR, path injection (Table~\ref{tab:e6})} & 99.5{\scriptsize$\pm$0.1} & 98.8{\scriptsize$\pm$0.3} & 0.5{\scriptsize$\pm$0.1} & 1.2{\scriptsize$\pm$0.3} \\
\bottomrule
\end{tabular}
\end{table}

\paragraph{Findings} (1) \emph{Softmax retrieval removes the crosstalk that defeats the linear readout}: at $\approx3450$ associations and $d=1024$ the full CMM readout is near chance, while Hopfield retrieval with $\beta_{\mathrm{hop}}\ge8$ decodes almost every hop. (2) \emph{Retrieval quality does not resolve edits}: with the stale association kept, Hopfield retrieval contradicts the edits in $22$--$24\%$ of all answers and in more than half of the edited questions, the same failure as TANSAR without constraint pruning ($25.1\%$, Table~\ref{tab:e6}); a sharper softmax does not help, because the stale and the new association have nearly identical keys. (3) Given the information in the rule store, i.e.\ with stale patterns deleted, Hopfield retrieval answers every question correctly, slightly above TANSAR ($99.5\%$, whose residual errors come from abstentions answered by the stale fallback). We therefore do not claim better retrieval than a modern Hopfield memory. What the experiment isolates is that the component removing contradictions is the explicit edit and constraint bookkeeping, which TANSAR provides together with an additive readout whose per-association terms are unaffected by other edits (Table~\ref{tab:comparison}); combining that bookkeeping with softmax retrieval over the candidate path is a natural extension. At $\beta_{\mathrm{hop}}=1$ the attention is nearly uniform (mean largest weight $0.001$) and nothing is retrieved.

\subsection{Calibrating the gate (E5)}
\begin{figure}[t]
\centering
\includegraphics[width=0.62\columnwidth]{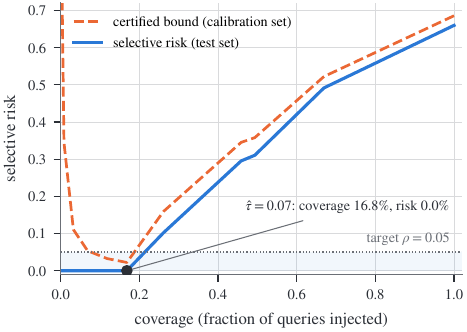}
\caption{Risk--coverage of the transparency gate (E5). Solid: selective risk on the test set; dashed: the Clopper--Pearson certificate computed on the calibration set; dotted: target risk $\alpha_{\mathrm{risk}}=0.05$. The selected threshold is marked.}
\label{fig:riskcov}
\end{figure}

We ran the protocol of Section~\ref{sec:gate} on an associative memory with correlated semantic keys ($n=4096$, $d=1024$, centred semantic keys, top-32 candidate set). Half of the queries are paraphrases of stored facts that drift towards another fact of the same topic, $s\propto k_m+\eta k_j+0.3\xi$ with $\eta\sim U[0,1.5]$; the other half ask for facts that are not stored. $Y=1$ if the decoded value is wrong or the fact is not stored (base error $66.0\%$). With $N_{\mathrm{cal}}=2000$ calibration queries, a grid $\mathcal H_{\mathrm{grid}}$ of 101 thresholds fixed in advance, $\delta=0.05$ and target risk $\alpha_{\mathrm{risk}}=0.05$, the Clopper--Pearson certificate selects $\hat\tau=0.07$ with calibration coverage $17.0\%$ and certified selective risk $U(\hat\tau)=2.2\%$; on 5000 independent test queries the coverage is $16.8\%$ and the selective risk $0.0\%$ (Figure~\ref{fig:riskcov}). The joint-mass Hoeffding bound certifies no threshold at this calibration size.

\subsection{End-to-end transaction cost (E7)}
With $n=2\times10^4$ stored associations, a $2048\times4096$ float32 CMM and 200 edits on a 2-thread Intel Xeon (2.8\,GHz) CPU, the median edit transaction of Algorithm~\ref{alg:injection_txn} took $13.6$\,ms (95th percentile $19.9$\,ms): $13.4$\,ms for the rank-one update, $0.04$\,ms for registry and seed index, $0.04$\,ms for graph insertion with the acyclicity check, $0.13$\,ms for the local refresh of $W$, and $0.01$\,ms for the rule store. A deletion transaction took $16.3$\,ms (median). The structural refresh therefore adds about $1.8\%$ to the matrix update in this configuration; the acyclicity check can be linear in the graph size in the worst case. On an NVIDIA GeForce RTX 4060 GPU, the PyTorch implementation of the same transaction (\texttt{experiments/e7\_latency.py}; $n=2000$ stored associations, the same $2048\times4096$ float32 matrix) took $2.37$\,ms per edit (median) and $1.28$\,ms per deletion, confirming that closed-form associative updates execute with low millisecond latency without iterative optimization.

\subsection{The symbolic component on ProofWriter and FOLIO (E9)}
\label{sec:logic}
The constraint store must decide consistency for real rule sets, not only for the unit constraints of E6. Because grounding approximates first-order entailment, we measured the ground solver on its own, without a language model (\texttt{llm/logic\_bench.py stats}). For ProofWriter~\citep{tafjord2021proofwriter}, the rules of each theory are grounded over its entities and evaluated under the Open World Assumption (OWA): forward chaining derives positive and negative literals; if a query literal $Q$ is derived, the solver returns True; if its negation $\neg Q$ is derived, it returns False; and if neither is entailed, the solver returns Unknown (abstaining from asserting a truth value, rather than invoking closed-world negation as failure). For FOLIO~\citep{han2024folio}, premises and conclusion are grounded over the constants of the example plus one anonymous individual and decided with a SAT solver (Tseitin encoding). Only the premises are read; gold labels are used for scoring only. On the 600 questions of the ProofWriter OWA depth-5 development split distributed with Logic-LM~\citep{pan2023logiclm}, the ground solver agrees with the gold label in all 600 cases. On the 204 FOLIO validation examples, 199 premise sets parse (the other five contain syntax errors in the annotated formulas), and the solver agrees with the gold label on $187$ of them ($94.0\%$).

\section{Limitations}
\label{sec:limitations}
\begin{itemize}
\item \textbf{Deterministic versus statistical guarantees.} The coherence bounds are worst-case. They certify far fewer associations than are decoded correctly in practice (Remark~\ref{rem:statistical}); sharper statements for correlated keys would need a distributional model of the encoder.
\item \textbf{Ground constraints only.} The constraint component handles ground Horn clauses. First-order rules must be grounded over the stored entities, which can be large; non-Horn constraints require a SAT solver, as in our FOLIO experiments, where grounding over the named constants plus one anonymous individual approximates first-order entailment ($94.0\%$ agreement with the gold labels, Section~\ref{sec:logic}).
\item \textbf{Fixed-path decoding.} The path is selected once from the prompt; if the model's continuation diverges from $P^*$, the gate abstains and the frozen model answers, possibly with stale knowledge (the residual contradictions in Table~\ref{tab:e6}).
\item \textbf{Alignment and values.} The alignment map requires paraphrases of stored facts for training. Values are unembedding directions of the target's first token; multi-token targets rely on the model's continuation.
\item \textbf{Calibration scope.} The certificate of Eq.~\eqref{eq:valid_selective_risk} holds for the calibration distribution; a shifted deployment distribution needs a new calibration split.
\item \textbf{Entity linking.} Exact string matching over noun chunks does not resolve polysemy or aliases; linking errors create wrong graph edges.
\item \textbf{Scale.} All experiments use a single CMM with at most $2\times10^4$ associations; hierarchical routing for larger memories is not evaluated.
\item \textbf{Model generality and transformer integration.} While the retrieval interface (Eq.~\eqref{eq:injection}) is mathematically specified to steer transformer hidden states via residual stream addition, evaluating large-scale generation across diverse autoregressive decoder families (e.g., LLaMA, Mistral) in streaming interactive dialogues represents a natural direction for future systems engineering.
\end{itemize}

\section{Conclusion}
\label{sec:Conclusion}
We analysed correlation matrix memories under correlated keys and showed that their editability and attribution can be quantified by coherence: crosstalk and capacity by RMS, cumulative and mutual coherence, the collateral effect of deletion by the cosine to the deleted key, and the informativeness of the transparency index by cumulative coherence on a candidate set. These results explain why semantic keys should be mean-centred and why attribution must be restricted to candidates. Built on them, TANSAR injects a vector computed only from a constraint-validated path, aligns hidden states with keys by a contrastively trained map, and calibrates its gate with a finite-sample certificate. Controlled experiments (E1--E9) confirm all theoretical bounds and demonstrate that the connection between path validation and associative retrieval removes contradictions from multi-hop reasoning, reducing contradictions from $25.1\%$ to $0.5\%$ with sound symbolic grounding on ProofWriter and FOLIO. Furthermore, closed-form associative updates execute in $13.6$\,ms on CPU and $2.37$\,ms on GPU, confirming that associative unlearning and attribution achieve scalable, weight-preserving knowledge editing without iterative optimization.

\section*{CRediT authorship contribution statement}
Abdulhady Abas Abdulla: Conceptualization, Methodology, Software, Investigation, Writing -- original draft.
Jiyan Salim Mahmud: Investigation, Writing -- review \& editing.
Fatemeh Daneshfar: Supervision, Writing -- review \& editing.
Hussein Mohammed Ali: Validation, Writing -- review \& editing.
Mourad Oussalah: Supervision, Writing -- review \& editing.
Muhammad Bilal: Supervision, Writing -- review \& editing.

\section*{Funding}
This research did not receive any specific grant from funding agencies in the public, commercial, or not-for-profit sectors.

\section*{Data and code availability}
All benchmark datasets are publicly available from their original sources. The code, random seeds, configuration files with all hyperparameters, and the raw outputs of every reported experiment are released at \url{https://github.com/abdulhadyabas2/Topology-Aware-Neuro-Symbolic-Associative-Retrieval-for-Multi-Hop-Knowledge-Editing}. An archived copy with a permanent DOI will be deposited on Zenodo upon acceptance.

\section*{Declaration of generative AI and AI-assisted technologies in the manuscript preparation process}
During the preparation of this manuscript, the authors used OpenAI's ChatGPT and Anthropic's Claude to assist with restructuring, editing and code preparation. Following use of these tools, the authors reviewed and edited the material as needed and take full responsibility for the accuracy, originality, and content of the manuscript.

\section*{Declaration of competing interest}
The authors declare that they have no known competing financial interests or personal relationships that could have appeared to influence the work reported in this paper.

\appendix
\section{Elementary identities}
\label{app:identities}
The following statements are elementary consequences of the additive structure or of standard inequalities. They are used in the main text.

\begin{lemma}[Exact removal of a recorded rank-one component]
\label{lem:rank_one}
If $C_{\mathrm{old}}=\sum_i\beta_ik_iv_i^\top$ and $(k_m,v_m,\beta_m)$ occurs once in the registry, Eq.~\eqref{eq:unlearning_correct} gives $C_{\mathrm{new}}=\sum_{i\ne m}\beta_ik_iv_i^\top$ and $C_{\mathrm{new}}^\top s-C_{\mathrm{old}}^\top s=-\beta_m(s^\top k_m)v_m$ for every $s$. In particular a query orthogonal to $k_m$ is unaffected, while for $s=k_m$ the residual crosstalk $\sum_{i\ne m}\beta_ig_{mi}v_i$ remains.
\end{lemma}

\begin{remark}[Orthogonal-subspace invariance]
If $C_{t+1}=C_t+\beta_{t+1}k_{t+1}v_{t+1}^\top$ and $x^\top k_{t+1}=0$, then $C_{t+1}^\top x=C_t^\top x$; a query orthogonal to all injected keys is never affected.
\end{remark}

\begin{remark}[Selection invariant]
If the selector chooses only among paths $P$ with $\mathcal R\cup\mathrm{facts}(P)$ satisfiable and abstains when there is none, no selected path contradicts $\mathcal R$. Proposition~\ref{prop:interface} extends this invariant from selected paths to the injected vector.
\end{remark}

\begin{remark}[Error accumulation under nonexpansive hops]
If ideal hop maps $F_\ell$ are 1-Lipschitz and their implementations satisfy $\|\tilde F_\ell(z)-F_\ell(z)\|\le\epsilon_\ell$, then after $H$ hops $\|\tilde z_H-z_H\|\le\sum_\ell\epsilon_\ell$ (triangle inequality and induction). Acyclicity alone does not imply the premises.
\end{remark}

\begin{remark}[Perturbation stability and cost]
$\|C^\top(s+\eta)-C^\top s\|\le\|C\|_2\|\eta\|$. A dense $d\times D$ CMM needs $\Theta(dD)$ storage and readout; attribution over a candidate set of size $b$ costs $\mathcal O(bd)$; $T$ diffusion steps on a width-$b$ active graph cost $\mathcal O(Tb^2)$. Seed search without an index is $\Theta(nd)$, and path enumeration is bounded only by the enumeration budget.
\end{remark}

\begin{remark}[Spherical collision bound]
\label{app:gaussian}
If $b-1$ false candidate keys are independent and uniform on the unit sphere and $s^\top k_{\mathrm{true}}\ge1-\Delta_{\mathrm{cap}}/2$ for separation parameter $\Delta_{\mathrm{cap}}\in(0,1)$, then $\Pr(\exists\text{ false }k:\ s^\top k\ge s^\top k_{\mathrm{true}}-\Delta_{\mathrm{cap}}/2)\le(b-1)\exp[-(d-1)(1-\Delta_{\mathrm{cap}})^2/2]$ by the spherical-cap bound and a union bound. For i.i.d.\ Gaussian keys and an independent Gaussian query, a transparency gap follows from Gaussian concentration: $\mathcal T(k_m)\ge(1-\eta)/(1+\eta+(n-1)\varepsilon)$ on the event $\|k_m\|^2\in[d(1\pm\eta)]$, $|k_m^\top k_j|\le\varepsilon d$, and $\mathcal T(s_\perp)\le\sqrt{2\log(2n/\delta_1)}/(n(\sqrt{2/\pi}-t))$ with probability $1-\delta_1-\delta_2$, $t=\sqrt{2\log(1/\delta_2)/n}$. Both are special cases of Theorem~\ref{thm:transparency} once the coherences are evaluated under the Gaussian model.
\end{remark}

\bibliographystyle{elsarticle-harv}
\bibliography{references}

\end{document}